# NONLINEAR VECTOR FILTERING FOR IMPULSIVE NOISE REMOVAL FROM COLOR IMAGES

M. Emre Celebi[1], Hassan A. Kingravi[2], Y. Alp Aslandogan[3]

ecelebi@lsus.edu, kingravi@cc.gatech.edu, yaaslandogan@pvamu.edu

[1]Department of Computer Science

Louisiana State University in Shreveport, Shreveport, LA 71115, USA

[2]Department of Computer Science

Georgia Institute of Technology, Atlanta, GA 30332, USA

[3]Department of Computer Science

Prairie View A&M University, Prairie View, TX 77446, USA

## ABSTRACT

In this paper, a comprehensive survey of 48 filters for impulsive noise removal from color images is presented. The filters are formulated using a uniform notation and categorized into 8 families. The performance of these filters is compared on a large set of images that cover a variety of domains using three effectiveness and one efficiency criteria. In order to ensure a fair efficiency comparison, a fast and accurate approximation for the inverse cosine function is introduced. In addition, commonly used distance measures (Minkowski, angular, and directional-distance) are analyzed and evaluated. Finally, suggestions are provided on how to choose a filter given certain requirements.

## I. INTRODUCTION

The growing use of color images in diverse applications such as medical image analysis, content-based image retrieval, remote sensing, and visual quality inspection has led to an increasing interest in color image processing. These applications involve many of the same tasks as their grayscale counterparts, such as edge detection, segmentation and feature extraction [1]. However, color images are often contaminated with noise which not only lowers their visual quality, but also complicates automated processing. Therefore, the removal of such noise is often a necessary preprocessing step for color image processing applications [2].

Image noise can come from many sources, and can be introduced into an image either during acquisition or transmission through sensors or communication channels, respectively [3]. ‘Impulsive noise’ is noise of low duration and high energy that can be caused either by faulty sensors or by electrical disturbances such as lightning and the operation of high-voltage machinery corrupting the transmission signal [4]. The introduction of such noise into an image is often detrimental to its future usage. If the image is meant for human consumption, the presence of noise lowers its perceptual quality. On the other hand, if it is to be processed further, the noise can make complex tasks such edge detection and segmentation even more difficult.

Numerous filters have been proposed in the literature for impulsive noise removal from color images. Among these, nonlinear filters have proved successful in the preservation of edges and fine image details while removing the noise [5]. The early approaches to nonlinear filtering of color images often involved the application of a scalar filter to each color channel independently. However, since separate processing ignores the inherent correlation between the color channels, these methods often introduce color artifacts to which the human visual system is very sensitive [6]. Therefore, vector filtering techniques that treat the color image as a vector field and process color pixels as vectors are more appropriate [7]. An important class of nonlinear vector filters is the one based on robust order-statistics with the vector median filter (*VMF*) [8] being the most widely known example. These filters involve reduced ordering [9][10] of a set of input vectors within a window to compute the output vector. Recent applications of these include enhancement of cDNA microarray images [11][12], virtual restoration of artworks [13][14], and video filtering [15][16][17][18].

The motivation of this study is two-fold. First, a large number of nonlinear vector filters have been proposed in the literature since 1990. Therefore, a study that categorizes and presents these filters in a unified notation is desirable. Second, to the best of the authors' knowledge, no study to date has objectively compared the performance of these filters on a large and diverse set of images. A similar study [19] presents a detailed survey of the nonlinear vector filters, noise models, filtering performance criteria, and applications; however it does not provide an experimental comparison of these filters.

In this study, 48 impulsive noise removal filters are presented in a systematic fashion and categorized into 8 families. Furthermore, the performance of these filters in terms of both effectiveness and efficiency are compared on a set of 100 images that cover a multitude of domains. In order to ensure fairness in the efficiency comparisons, a fast and accurate approximation for the inverse cosine function (used in many of the filters) is introduced. In addition, the relative merits of commonly used distance measures (Minkowski, angular, and directional-distance) are analyzed and compared. Finally, suggestions are provided on how to choose a filter given certain requirements.

The rest of the paper is organized as follows. Section 2 introduces the notation and categorizes the filters. Section 3 describes the image set, the noise models, and the filtering performance criteria. Finally, section 4 discusses the experimental results and gives the conclusions.

## II. CATEGORIZATION OF THE FILTERS

In this section, the 48 impulsive noise removal filters are categorized into 8 groups as follows:

1. Basic Vector Filters
2. Adaptive Fuzzy Vector Filters
3. Hybrid Vector Filters
4. Adaptive Center-Weighted Vector Filters
5. Entropy Vector Filters
6. Peer Group Vector Filters
7. Vector Sigma Filters
8. Miscellaneous Vector Filters

The notation used in the descriptions of these filters is shown in Table 1. Note that the author recommended parameter values for each filter are indicated in the descriptions.

**INSERT TABLE 1 HERE**

### 2.1 Basic Vector Filters

These are the earliest impulsive noise removal filters proposed in the literature. The subsequent, more advanced filters are more or less based on these basic filters. Table 2 shows the mathematical expressions for these filters.

#### 2.1.1 Vector Median Filter

The Vector Median Filter (*VMF*) [8] and its extensions [20][21] follows directly from the concept of the nonlinear order statistics in that the output of the filter is the lowest ranked vector in the window. The *VMF* orders the color input vectors according to their relative magnitude differences using the Minkowski metric as a distance measure. The two most widely used such measures are the *L1* (Manhattan distance) and the *L2* (Euclidean distance) norms [22].

#### 2.1.2 Alpha-Trimmed Vector Median Filter

The Alpha-Trimmed Vector Median Filter (*ATVMF*) [18] selects the lowest ranked $1 + \alpha$ vectors as input to an averaging filter. The trimming operation guarantees good performance in the presence of impulsive noise. In addition, the averaging operation helps the filter cope with Gaussian noise. The parameter $\alpha$ is set to $\lfloor n/2 \rfloor$.

#### 2.1.3 Basic Vector Directional Filter

Another method for detecting the outliers in a window is to rank the color vectors based on the orientation difference between them. In other words, vectors with atypical directions are considered to be outliers. The Basic Vector Directional Filter (*BVDF*) [23] uses this concept in a manner similar to the *VMF*, by using the angle between two color vectors as the distance criterion. Since the direction of a vector corresponds to its chromaticity [24], this filter preserves the chromaticity of the input vectors better than the *VMF*.

### 2.1.4 Generalized Vector Directional Filter

The Generalized Vector Directional Filter (*GVDF*) [24] is a generalization of the *BVDF* in that its output is a superset of the single *BVDF* output. After the vectors are ranked according to the angular distance criterion, a set of low rank vectors are selected as input to an additional filter to produce a single output vector. In the second step, only the magnitudes of the vectors are considered. Thus, any gray-scale filter [25] such as the arithmetic mean filter (*AMF*), the multistage median filter, and various morphological filters can be used. In this study, the *AMF* is used for magnitude processing.

### 2.1.5 Directional Distance Filter

The Directional Distance Filter (*DDF*) [26][27] is a combination of the *VMF* and the *BVDF* derived by the simultaneous minimization of their defining functions (see Table 2). The motivation behind this is to incorporate information about both a vector's magnitude (brightness) and its direction (chromaticity) in the calculation of the distance metric. The parameter $\gamma$ in this case controls the relative importance of each component. This parameter is set to 0.5, which implies an equal consideration for both measures.

### 2.1.6 Content Based Rank Filter

The Content Based Rank Filter (*CBRF*) [28], like the *DDF*, ranks the vectors according to a distance metric that incorporates more information about the vector as a whole than the criterions used by the *VMF* and the *BVDF*. The similarity between two vectors in this case can be expressed as the ratio of some function of what they share (commonality) to what they comprise (totality) [29]. The numerator (commonality) and the denominator (totality) correspond to the vector difference and the vector sum, respectively.

**INSERT TABLE 2 HERE**

## 2.2 Adaptive Fuzzy Vector Filters

These filters utilize data-dependent coefficients to adapt to local image characteristics [11][30][31]. The general form of an adaptive fuzzy vector filter is given as a nonlinear transformation of a fuzzy weighted average of the input vectors within a window *W*:

$$x_{afvf} = g\left(\frac{\sum_{i=1}^{n} w_i x_i}{\sum_{i=1}^{n} w_i}\right) \quad (1)$$

where $x_{afvf}$ is the filter output, $g(.)$ is a nonlinear function, and $w_i \geq 0 \Leftrightarrow x_i$ are the fuzzy weights that correspond to each input vector. The weights provide the degree to which an input vector contributes to the filter output and are determined by fuzzy transformations of the cumulative distances associated with each input vector.

2.2.1 Fuzzy Weighted Average Filters

In the Fuzzy Weighted Average Filters (*FWAFs*) the function $g(.)$ is the identity function:

$$x_{fwaf} = \frac{\sum_{i=1}^{n} w_i x_i}{\sum_{i=1}^{n} w_i} \quad (2)$$

Because of the averaging operation, the filter output $x_{fwaf}$ is generally not included in the input vector set $\{x_1, x_2, \ldots, x_n\}$. This allows better performance in the presence of Gaussian noise when compared to pure order-statistics based filters that select the output vector from the set of input vectors. Note that depending on the distance criterion and the corresponding fuzzy transformation, various fuzzy filters can be derived from (2).

2.2.1.1 Fuzzy Vector Median Filter

In the Fuzzy Vector Median Filter (*FVMF*) [30][31][32] the Minkowski metric is used as the distance function and the fuzzy membership function has an exponential form. In this case the fuzzy weights are given by:

$$w_i = \exp\left(-l^{\gamma}(i)/\beta\right) \quad for \quad i = 1, 2, \ldots, n \qquad (3)$$

where $\gamma$ and $\beta$ are parameters that control the amount of fuzziness in the weights [33]. The following values are used for these parameters: $\gamma = 0.5$ and $\beta = 1.0$.

2.2.1.2 Fuzzy Vector Directional Filter

In the Fuzzy Vector Directional Filter (*FVDF*) [30][31] the vector angle metric is used as the distance function and the fuzzy membership function has a sigmoidal form. In this case the fuzzy weights are given by:

$$w_i = \frac{\beta}{\left(1 + \exp\left(a(i)\right)\right)^{\gamma}} \quad for \quad i = 1, 2, \ldots, n \qquad (4)$$

where $\gamma$ is a parameter that can be used to adjust the weighting effect of the membership function and $\beta$ is a weight-scale threshold. The following values are used for these parameters: $\gamma = 1.0$ and $\beta = 2.0$.

2.2.1.3 Adaptive Nearest Neighbor Filter

In the Adaptive Nearest Neighbor Filter (*ANNF*) [34] the fuzzy weights are determined as follows:

$$w_i = \frac{a_{(n)} - a_{(i)}}{a_{(n)} - a_{(1)}} \quad for \quad i = 1, 2, \ldots, n \qquad (5)$$

where $a_{(n)}$ and $a_{(1)}$ are the maximum and minimum cumulative angular distances, respectively. It should be noted that other distance measures such as the Minkowski and directional-distance functions can also be used in (5).

2.2.1.4 Adaptive Nearest Neighbor Multichannel Filter

The Adaptive Nearest Neighbor Multichannel Filter (*ANNMF*) [35] is a modification of the *ANNF* that uses a composite distance function rather than an angular one:

$$D(x_i, x_j) = 1 - \left( \frac{< x_i, x_j >}{\|x_i\| \cdot \|x_j\|} \right) \left( 1 - \frac{\left| \|x_i\| - \|x_j\| \right|}{max\left( \|x_i\|, \|x_j\| \right)} \right) \quad (6)$$

2.2.2 Fuzzy Ordered Vector Filters

The Fuzzy Ordered Vector Filters [31][36] are a fuzzy generalization of the alpha-trimmed filters in which the input vectors are ordered according to their fuzzy membership strengths and only those vectors with the largest fuzzy weights contribute to the output vector:

$$x_{FOVF} = \frac{\sum_{i=1}^{k} w_{(i)} x_{(i)}}{\sum_{i=1}^{k} w_{(i)}} \quad , \quad k \in [1, n] \quad (7)$$

where $x_{(k)} \leq x_{(k-1)} \leq \cdots \leq x_{(1)}$ are the vectors with the $k$ largest weights $w_{(k)} \leq w_{(k-1)} \leq \cdots \leq w_{(1)}$, respectively.

The number of vectors ($k$) can be determined adaptively by considering only those input vectors with fuzzy weights greater than $1/n$ [30]. Note that any fuzzy membership function such as (3), (4), or (5) can be used to determine the weights in (7). In this study, only the Fuzzy Ordered Vector Median Filter (*FOVMF*) {Equations (3) and (7)} and the Fuzzy Ordered Vector Directional Filter (*FOVDF*) {Equations (4) and (7)} are considered.

## 2.3 Hybrid Vector Filters

These filters utilize a number of sub-filters of different types (hence the term 'hybrid') and define the output as a linear or nonlinear combination of the input vectors [37]. Consequently, the output is often not included in the input set. Table 3 shows the mathematical expressions for these filters.

### 2.3.1 Extended Vector Median Filter

The Extended Vector Median Filter (*EXVMF*) [8] combines the *VMF* with linear filtering to compensate for the deficiency of the *VMF* in dealing with Gaussian noise. Near edges this filter behaves like the *VMF* and preserves the details, while in smooth areas it behaves like the *AMF*, resulting in improved noise attenuation.

### 2.3.2 Hybrid Directional Filter

The Hybrid Directional Filter (*HDF*) [38] is also based on the concept of independent vectorial attribute processing introduced in the *DDF*. It can be thought of as a nonlinear combination of the *VMF* and the *BVDF* filters.

### 2.3.3 Adaptive Hybrid Directional Filter

The Adaptive Hybrid Directional Filter (*AHDF*) [38] is an extension of the *HDF* which utilizes the *AMF* in the filter structure. This is so that the magnitude of the output vector will be that of the mean vector in smooth regions and that of the median operator near edges. Note that the criteria for the selection of the output vector in this filter is similar to the one used in the *EXVMF*.

### 2.3.4 Vector Median-Rational Hybrid Filter

The Vector Median-Rational Hybrid Filter (*VMRHF*) [39][40][41] is a multichannel extension of the median-rational hybrid filter that combines the output of three sub-filters (two vector median filters and a center weighted vector median filter[♠]) in a rational function. It differs from a linear low-pass filter mainly due to the scaling which is essentially an edge-sensing term characterized by the Euclidean distance between the two *VMF* outputs. The coefficient vector $\alpha = [\alpha_1\ \alpha_2\ \alpha_3]$ in the numerator is chosen apriori and serves to weight the outputs of the three sub-filters. The parameters $\beta_1$ and $\beta_2$ in the denominator are positive constants. The former ensures numerical stability while the latter regulates the nonlinearity. The masks utilized by each sub-filter are as follows:

[♠] see section 2.4

$$VMF_1 : \begin{bmatrix} 0 & 1 & 0 \\ 1 & 1 & 1 \\ 0 & 1 & 0 \end{bmatrix} \quad , \quad CWVMF : \begin{bmatrix} 1 & 1 & 1 \\ 1 & 3 & 1 \\ 1 & 1 & 1 \end{bmatrix} \quad , \quad VMF_2 : \begin{bmatrix} 1 & 0 & 1 \\ 0 & 1 & 0 \\ 1 & 0 & 1 \end{bmatrix} \qquad (8)$$

Note that only those pixels with non-zero coefficients are considered in each of these masks. The parameter values are chosen as follows: $\alpha_1 = 1.0$, $\alpha_2 = -2.0$, $\alpha_3 = 1.0$, $\beta_1 = 3.0$, and $\beta_2 = 3.0$.

2.3.5 Fuzzy Rational Hybrid Filters

The Fuzzy Rational Hybrid Filters [36][42][43] are a family of adaptive hybrid filters that are derived from the *VMRHF*. In the Fuzzy Vector Median-Rational Hybrid Filter (*FVMRHF*) one of the sub-filters is a fuzzy center-weighted vector median filter (*FCWVMF*) and the other two are fuzzy vector median filters (*FVMF*). The fuzzy weights for these sub-filters are given by:

$$w_i = \frac{2}{1 + \exp\left(l^{\gamma}(i)\right)} \quad for \;\; i = 1, 2, \ldots, n \qquad (9)$$

The Fuzzy Vector Directional-Rational Hybrid Filter (*FVDRHF*) and the Fuzzy Directional Distance-Rational Hybrid Filter (*FDDRHF*) are the angular and the directional-distance counterparts of the *FVMRHF*, respectively. The smoothing parameter $\gamma$ is set to 1.0, and for the remaining parameters the *VMRHF* values are used.

2.3.6 Kernel Vector Median Filter

The Kernel Vector Median Filter (*KVMF*) [44][45][46][47][48] outputs a vector that lies somewhere between the center pixel and the *VMF* output. In other words, the output vector is a linear combination of the two vectors. The weights are determined by the kernel $\mu$ for which several choices such as Laplacian, Gaussian, Cauchy, Epanechnikov, etc. are available. Table 3 gives the filter formulation for the Laplacian kernel with the normalization factor $\beta$ and the kernel width $h$. The value of $\beta$ depends on the kernel of choice ($\beta$ = 0.5 for the Laplacian kernel). The parameter $h$ can be estimated from the entire image as shown in Table 3.

The operation of this filter represents a compromise between the *VMF* and the identity operation. The kernel is a function of the distance between the center pixel and the *VMF* output; if the center pixel is not noisy, then the kernel function is close to 1, and the output will be close to the original value of the center pixel. Otherwise, the output will be close to the *VMF* output.

**INSERT TABLE 3 HERE**

2.4 Adaptive Center-Weighted Vector Filters

The vector median filter can be generalized by associating with each pixel $x_i$ a non-negative integer-valued weight [18][49]:

$$x_{WVMF} = \underset{x_i \in W}{argmin}\left( \sum_{j=1}^{n} w_j \left\| x_i - x_j \right\| \right) \quad (10)$$

This filter is called the Weighted Vector Median Filter (*WVMF*). Note that by replacing the distance function in (10) with the angular or directional-distance functions, one can obtain the analogous Weighted Vector Directional Filter (*WVDF*) or Weighted Directional-Distance Filter (*WDDF*), respectively [50][51].

The flexible form of the Weighted Vector Filters allows one to design an optimal filter for a particular domain by adjusting the weights. The weights are often determined by an optimization procedure using a number of training images [50][52][53][54]. If only the center weight is varied while the others are fixed, the *WVMF* simplifies to the Center-Weighted Vector Median Filter (*CWVMF*) [55][56]:

$$
\begin{aligned}
x_{CWVMF^k} &= \underset{x_i \in W}{argmin}\left(\sum_{j=1}^{n} w_j(k) \cdot \left\| x_i - x_j \right\|\right) \\
w_j(k) &= \begin{cases} n - 2k + 2 & for \;\; j = C \\ 1 & otherwise \end{cases} \quad , \quad k \in [1, C]
\end{aligned}
\tag{11}
$$

When the smoothing parameter $k = 1$, the *CWVMF* is equivalent to the identity filter and thus no smoothing is performed. As the value of $k$ is increased, the smoothing capability of the filter increases. Finally, when $k$ attains its maximum value $C$, the filter becomes equivalent to the *VMF*, and the maximum amount of smoothing is performed. Similar formulations can be derived for the angular and directional-distance functions.

### 2.4.1 Adaptive Center-Weighted Vector Filters

The Adaptive Center-Weighted Vector Filters [55][57], i.e. *ACWVMF*, *ACWVDF*, and *ACWDDF*, employ a user-specified threshold to determine whether the center pixel is noisy or not. If the center pixel is noisy, it is replaced by the output of one of the three basic vector order-statistics filters, the *VMF*, the *BVDF*, or the *DDF*. Otherwise, it remains unchanged. The mathematical expressions for these filters are given in Table 4. The thresholds are set to 80, 0.19, and 10.8 for the *ACWVMF*, *ACWVDF*, and *ACWDDF*, respectively. The $\lambda$ parameter is set to 2.

An alternative design for the adaptive center-weighted filters is proposed in [58]. Extensions of these filters for image sequence processing and efficient hardware implementations can be found in [15][17].

2.4.2 Modified Center-Weighted Vector Median Filter

The Modified Center-Weighted Vector Median Filter (*MCWVMF*) [59][60] is a modification of the *CWVMF* in which only the cumulative distance associated with the center pixel is weighted. In contrast, in the *CWVMF* the center weight contributes to all of the cumulative distance values except for that associated with the center pixel. This allows the *MCWVMF* to be faster than the *CWVMF* since fewer multiplications are involved in the former. Table 4 shows the mathematical expression of the *MCWVMF*. Note that the center weight $w$ in the *MCWVMF* is a real number between 0 and 1, whereas the one in the *CWVMF* is a nonnegative integer. The $w$ parameter is set to 0.5.

**INSERT TABLE 4 HERE**

## 2.5 Entropy Vector Filters

Entropy Vector Filters [61][62] are a family of adaptive switching filters that are multichannel extensions of the gray-scale local contrast entropy filter [63]. For the gray-scale case, the contrast of a pixel $x_i$ within a window $W$ can be expressed as:

$$C_i = \frac{|x_i - \overline{x}|}{\overline{x}} = \frac{\Delta_i}{\overline{x}} \quad (12)$$

where $\overline{x}$ denotes the mean gray-level. The local contrast probability $P_i$ and local contrast entropy $H_i$ associated with pixel $x_i$ are given by:

$$P_i = \frac{\Delta_i}{\sum_{j=1}^{n} \Delta_j} \qquad (13)$$
$$H_i = -P_i \, log P_i$$

Noisy pixels heavily contribute to the total local contrast entropy which is given by:

$$H = \sum_{i=1}^{n} H_i \quad (14)$$

Extensions of this formulation for the multichannel case are given in Table 5. These filters, i.e. *EVMF*, *EBVDF*, and *EDDF*, employ an adaptive threshold (the fraction of local contrast entropy contributed by the center pixel) to determine whether the center pixel is noisy or not. If the center pixel is noisy, it is replaced by the output of one of the three basic vector filters, the *VMF*, the *BVDF*, or the *DDF*. Otherwise, it remains unchanged. An extension of the entropy filters for color video sequence enhancement can be found in [64].

**INSERT TABLE 5 HERE**

2.6 Peer Group Vector Filters

These are adaptive switching filters based on the peer group concept [65]. Essentially, the peer group of a pixel in a given window represents the set of neighboring pixels that are sufficiently similar to it according to a particular measure. Table 6 shows the mathematical expressions for these filters.

2.6.1 Peer Group Filter

In the Peer Group Filter (*PGF*) [65] the pixels in the window are sorted in ascending order according to their distances to the center pixel. The peer group of the center pixel is then determined as the $m = (\sqrt{n} + 1)/2$ pixels that rank the lowest in this sorted sequence. Next, in order to remove the effect of the impulsive noise, the first order differences δ(i) are calculated. Finally, the center pixel is considered noisy if one of these difference values is greater than a user-specified threshold. In this case, the center pixel is replaced with the *VMF* output; otherwise it remains unchanged. The threshold $T$ is set to 45.

2.6.2 Fast Peer Group Filter

The Fast Peer Group Filter (*FPGF*) [4] is a fast modification of the *PGF* in which the center pixel is considered to be noise-free as soon as *m* pixels in the window are determined to be sufficiently similar to it. If *m* is low, and the amount of noise in the image is not very high, this allows for a dramatic reduction in the number of distance computations that need to be performed. The parameters *m* and *T* are set to 3 and 45, respectively.

**INSERT TABLE 6 HERE**

## 2.7 Vector Sigma Filters

Vector Sigma Filters [66][67][68][69][70] are a family of adaptive switching filters that are multichannel extensions of the gray-scale sigma filter [71]. These filters utilize approximations of the multivariate variance within a window to determine whether the center pixel is noisy or not. If the center pixel is noisy, it is replaced by the output of one of the three basic vector filters, the *VMF*, the *BVDF*, or the *DDF*. Otherwise, it remains unchanged.

The concept of variance can be extended to the multivariate case using the covariance matrix. Scalar measures for multivariate variance can be calculated from this matrix as the sum or product of the eigenvalues [72]. However, computing the variance within each window in this manner is computationally very expensive. Therefore, Vector Sigma

Filters employ approximations of the multivariate variance based on either the mean vector or the lowest ranked vector.

The members of the Vector Sigma Filter family are given in Table 7. The non-adaptive vector sigma filters (*SVMF*, *SBVDF*, and *SDDF*) require a tuning parameter $\lambda$ to determine the switching threshold, while the adaptive vector sigma filters (*ASVMF*, *ASBVDF*, and *ASDDF*) determine this threshold adaptively. The parameter $\lambda$ is set to 4.0.

**INSERT TABLE 7 HERE**

## 2.8 Miscellaneous Vector Filters

This section contains the filters that do not fit into any of the categories described earlier. Table 8 shows the mathematical expressions for these filters. Some of these have commonalities with certain filters in other categories. For example, the Adaptive Multichannel Non-Parametric Filters resemble the *KVMF* in that they are based on similarity rather than dissimilarity (distance). However, they are not included in the Hybrid Vector Filters category since they do not utilize multiple sub-filters of different types.

### 2.8.1 Vector Signal-Dependent Rank Order Mean Filter

The Vector Signal-Dependent Rank Order Mean Filter (*VSDROMF*) [73] is an extension of the gray-scale *SDROM* filter [74]. In this filter, the pixels in the window are first sorted according to their cumulative distances to all other pixels. The distances between the center pixel and each of the lowest ranked 4 (for the general case $\lfloor n/2 \rfloor$) pixels are then compared against increasing thresholds. If any of these distances exceeds its respective threshold, the center pixel is considered to be noisy, and is replaced by the lowest ranked pixel, i.e. the *VMF* output. Otherwise, the center pixel remains unchanged. The thresholds are set to 35, 40, 45, and 50.

2.8.2 Adaptive Multichannel Non-Parametric Filters

The Adaptive Multichannel Non-Parametric Filters (*AMNF*s) [75][76] approach the filtering problem from an estimation theoretic perspective. Specifically, these filters are based on non-parametric kernel density estimation [77]. The general form of the *AMNFs* is given in Table 8. Two possible choices for the kernel function are the multivariate exponential $K(z) = e^{-|z|}$ (*AMNFE*) and the multivariate gaussian $K(z) = e^{-0.5 z^T z}$ (*AMNFG*) functions. The *k* parameter in the kernel width calculation is set to 0.33.

2.8.3 Fast Modified Vector Median Filter

In the Fast Modified Vector Median Filter (*FMVMF*) [78][79], the center pixel is replaced with the window pixel that minimizes the cumulative distance to all others (excluding the center pixel), provided that the difference between the cumulative distance associated with the center pixel and the minimum cumulative distance is greater than a threshold. Otherwise, the center pixel remains unchanged. Note that this scheme privileges the center pixel since its cumulative distance calculations involve $n$ - 1 terms, whereas the calculations associated with the other pixels involve $n$ - 2 terms. The distance threshold parameter is set to 0.75.

### 2.8.4 Adaptive Vector Median Filter and Adaptive Basic Vector Directional Filter

In the Adaptive Vector Median Filter (*AVMF*) [80], the center pixel is considered to be noisy if the distance between itself and the mean of the lowest ranked $k$ vectors is greater than a threshold. In this case, the center pixel is replaced by the *VMF* output. Otherwise, it remains unchanged.

The Adaptive Basic Vector Directional Filter (*ABVDF*) [81] is the angular counterpart of the *AVMF*. The thresholds are set to 100 and 0.16 for the *AVMF* and *ABVDF*, respectively. The $k$ parameters are both set to $\lceil n/2 \rceil$.

### 2.8.5 Fast Fuzzy Noise Reduction Filter

In the Fast Fuzzy Noise Reduction Filter (*FFNRF*) [82][83], the center pixel is replaced with the window pixel that maximizes the cumulative similarity to all others excluding the center pixel. Note that this center exclusion scheme is the same as in the *FMVMF*. The similarity between two pixels is determined using a special fuzzy metric [84] (see

Table 8). An interesting property of this metric is that the value of each term in the product can be pre-computed as:

$$Q^{\alpha}(a,b) = \left( \frac{min(a,b)+K}{max(a,b)+K} \right)^{\alpha} \quad (15)$$

Using the pre-computed values, the fuzzy similarity between two pixels $x_i$ and $x_j$ can be computed as:

$$M^{\alpha}(x_i, x_j) = \prod_{k=1}^{3} Q^{\alpha}(x_i, x_j) \quad (16)$$

It's empirically demonstrated that the computation of the fuzzy metric $M$ using the pre-computed values is even faster than that of the $L_1$ norm. The $K$ and $\alpha$ parameters are set to 1024 and 3.5, respectively.

**INSERT TABLE 8 HERE**

## III. EXPERIMENTAL SETUP

In this section, the image set that will be used in the experiments is first described. The impulsive noise models that are used to artificially corrupt the images for evaluation purposes are then presented. Finally, the filtering performance criteria that will be considered in the comparisons are detailed.

### 3.1 Image Set Description

In order to compare the performance of the filters on a wide variety of images, a set of 100 high quality *RGB* images was collected from the Internet. These included images of people, animals, plants, buildings, aerial maps, man-made objects, natural scenery, paintings, sketches, as well scientific, biomedical, and synthetic images and test images commonly used in the color image processing literature. Figure 1 shows representative images from this set.

**INSERT FIGURE 1 HERE**

## 3.2 Noise Models

Various simplified color image noise models have been proposed in the literature [3][5][18]. In this study, the following two impulsive noise models are considered:

1. *Uncorrelated Impulsive Noise*

$$x^k = \begin{cases} r^k & \text{with probability } \varphi, \\ o^k & \text{with probability } 1-\varphi \end{cases}$$

where $o = \{o^1, o^2, o^3\}$ and $x = \{x^1, x^2, x^3\}$ represent the original and noisy color vectors, respectively, φ denotes the channel corruption probability, and $r = \{r^1, r^2, r^3\}$ is a random vector that represents the impulsive noise such that $r^k \in [0,10]$ or $r^k \in [245, 255]$ with equal probability.

2. *Correlated Impulsive Noise*

$$x = \begin{cases} o & \text{with probability } 1-\varphi, \\ \{r^1, o^2, o^3\} & \text{with probability } \varphi_1 \cdot \varphi, \\ \{o^1, r^2, o^3\} & \text{with probability } \varphi_2 \cdot \varphi, \\ \{o^1, o^2, r^3\} & \text{with probability } \varphi_3 \cdot \varphi, \\ \{r^1, r^2, r^3\} & \text{with probability } \left(1-(\varphi_1+\varphi_2+\varphi_3)\right) \cdot \varphi \end{cases}$$

where $\varphi$ is the sample corruption probability and $\varphi_1$, $\varphi_2$, and $\varphi_3$ are the channel corruption probabilities. In this study, the following values are used: $\varphi_1 = \varphi_2 = \varphi_3 = 0.25$.

In the following discussion, a particular combination of a noise model and a noise level such as ‘5% correlated noise’ will be referred to as a ‘noise configuration’.

### 3.3 Filtering Performance Criteria

In order to evaluate the performance of the filters, three effectiveness and one efficiency criteria are employed. The effectiveness criteria are [5]:

1. *Mean Absolute Error* (*MAE*)

$$MAE = \frac{1}{3 \cdot M \cdot N} \sum_{i=1}^{M} \sum_{j=1}^{N} \left[ | R(i,j) - \hat{R}(i,j) | + | G(i,j) - \hat{G}(i,j) | + | B(i,j) - \hat{B}(i,j) | \right] \quad (17)$$

where $M$ and $N$ represent the image dimensions, $\{R(i,j), G(i,j), B(i,j)\}$ and $\{\hat{R}(i,j), \hat{G}(i,j), \hat{B}(i,j)\}$ are the $RGB$ coordinates of the pixel $(i,j)$ in the original and the filtered images, respectively. $MAE$ is a measure of the detail preservation capability of a filter.

2. *Mean Squared Error (MSE)*

$$MSE = \frac{1}{3 \cdot M \cdot N} \sum_{i=1}^{M} \sum_{j=1}^{N} \left[ (R(i,j) - \hat{R}(i,j))^2 + (G(i,j) - \hat{G}(i,j))^2 + (B(i,j) - \hat{B}(i,j))^2 \right] \quad (18)$$

$MSE$ is a measure of the noise suppression capability of a filter.

3. *Normalized Color Distance (NCD)*

$$NCD = \frac{\sum_{i=1}^{M} \sum_{j=1}^{N} \sqrt{[(L^*_{ab}(i,j) - \hat{L}^*_{ab}(i,j))^2 + (a^*(i,j) - \hat{a}^*(i,j))^2 + (b^*(i,j) - \hat{b}^*(i,j))^2]}}{\sum_{i=1}^{M} \sum_{j=1}^{N} \sqrt{(L^{*\,2}_{ab}(i,j) + a^{*2}(i,j) + b^{*2}(i,j)}} \quad (19)$$

where $\{L^*_{ab}(i,j), a^*(i,j), b^*(i,j)\}$ and $\{\hat{L}^*_{ab}(i,j), \hat{a}^*(i,j), \hat{b}^*(i,j)\}$ are the *CIE-*$L^*a^*b^*$ coordinates of the pixel $(i,j)$ in the original and the filtered images, respectively. $NCD$ is a perceptually oriented metric that measures the color preservation capability of a filter.

The efficiency of a filter is measured by the execution time in seconds. In order to ensure a fair comparison, all of the filters were implemented in the same style in the C language and compiled with the *gcc* 3.4 compiler. The experiments were performed on an Intel Pentium D 2.66GHz machine.

An issue in the comparison of the execution times is the cost of the inverse cosine (*acos*) function that is utilized in the angular and directional-distance filters. Standard library implementations of this function are computationally very expensive, causing angular distance computations to be much slower than the Minkowski distance computations. For example, on a typical 512 x 512 image, the *VMF* takes about 0.36 seconds, while the *BVDF* takes approximately 10.0 seconds. A solution to mitigate this problem is to use an approximation for the *acos* function over the interval [0, 1]. However, this is not easy because of the singularity of this function near 1. This can be circumvented using the following numerically more stable identity for $x \geq 0.5$ [85]:

$$\mathrm{acos}(x) = 2 \cdot \mathrm{asin}\left(\sqrt{(1-x)/2}\right) \quad (20)$$

In equation (20), the inverse sine (*asin*) function receives its arguments from the interval [0, 0.5]. Fortunately, this function is almost linear in this interval and can be accurately approximated using a third degree minimax polynomial [86]:

$$\mathrm{asin}(x) \cong \text{-0.67921302e-4+(1.003729762+(-0.309031329e-1+.2356774247} \cdot x) \cdot x) \cdot x \quad (21)$$

The approximation error is ε = 0.00006792131489. Similarly, the *acos* function is almost linear in the interval [0, 0.5] and can be approximated by:

$$\mathrm{acos}(x) \cong \text{1.570864248+(-1.003729768+(0.309031763e-1-.2356774861} \cdot x) \cdot x) \cdot x \quad (22)$$

The approximation error is ε = 0.00006792158693. This piecewise approximation of the *acos* function gives virtually the same numerical results, i.e. *MAE*, *MSE*, and *NCD*, as the standard *acos* function when used in the implementation of an angular or a directional-distance filter. This is because the error propagation is not very significant considering the small size of a filter window (only 9 pixels in a 3 x 3 window).

In order to demonstrate the effect of the approximation on the running time of an actual filter, the *BVDF* implementation that uses the standard *acos* function and the one that uses the approximation were both executed on the entire image set (100 images). The standard implementation took 1428 seconds, while the approximate one took 102 seconds. Similar gains in the execution times (≈ 13x-14x) were observed for the other angular and directional-distance filters as well. Note that methods used to speed up the *VMF* operation itself [87][88] are beyond the scope of this study.

## IV. DISCUSSION AND CONCLUSIONS

This section discusses the experimental results and presents the conclusions. First, the filters are compared based on the previously described measures of effectiveness and efficiency. Second, the filters that achieve a good compromise between effectiveness and efficiency are identified. Finally, the three commonly used distance measures are compared.

### 4.1 Discussion

Tables 9-12 show the rankings[♣] of the filters based on the following criteria: *MAE*, *MSE*, *NCD*, and execution time, respectively. The results[♦] are presented for the two noise models (uncorrelated and correlated impulsive noise) and three noise levels (5%, 10%, and 15%). The average rankings are obtained by averaging the individual filter rankings over the entire image set.

**INSERT TABLE 9 HERE**

**INSERT TABLE 10 HERE**

**INSERT TABLE 11 HERE**

**INSERT TABLE 12 HERE**

In order to determine the most effective filters at each noise level, we select the best 10 filters with respect to each effectiveness measure (*MAE*, *MSE*, *NCD*) for each noise model. Based on this selection, at each noise level, the filters that perform well regardless of the noise model and the effectiveness measure are determined (see Table 13). It can be observed that two filter families are particularly prominent in effectiveness: the Adaptive Center-Weighted Vector Filters and the Vector Sigma Filters. This can be attributed to the effectiveness of the noise detection criteria used in these families. By varying the smoothing parameter, the Adaptive Center-Weighted Vector Filters employ a

[♣] Note that the rankings start from 0 rather than 1.

[♦] For comparison purposes, the window size for each filter is set to 3 x 3 and the *L2* norm is used whenever the Minkowski distance is involved.

computationally expensive but robust iterative scheme to determine whether the center pixel is noisy or not. On the other hand, the Vector Sigma Filters utilize approximations of the multivariate variance within a window in their noise detection criteria. Interestingly, in general, the non-adaptive vector sigma filters perform better than their adaptive counterparts.

**INSERT TABLE 13 HERE**

The filters that are effective under any circumstances are those that appear in every row of Table 13. These are the *ACWDDF*, *PGF*, *SDDF_rank*, and *ACWVMF*. Among these filters, the *ACWDDF* consistently ranks the highest under different noise configurations. The *PGF* and *ACWVMF* have relatively stable rankings, whereas the *SDDF_rank* exhibits somewhat fluctuating behavior. Figures 2 and 3 show the results of these filters on two images corrupted by 10% and 15% correlated noise, respectively.

**INSERT FIGURE 2 HERE**

**INSERT FIGURE 3 HERE**

The execution time is also a very important factor that determines the practicality of a noise removal filter. As Table 12 shows, the ordering of the filters with respect to execution time remains almost unchanged across different noise configurations. The 10 most efficient filters are: *FPGF*, *ASVMF_mean*, *PGF*, *SVMF_mean*, *FFNRF*, *MCWVMF*,

*ASBVDF_mean*, *SVMF_rank*, *FMVMF*, and *VMF*. The following observations are in order:

- Except for the *VMF*, every filter in the list is based on the concept of switching (alternating between the identity and the filter operations).
- The *FPGF* is clearly the most efficient filter.
- The *PGF* is the only filter that ranks very high in terms of both effectiveness and efficiency. This is significant, considering the most effective filter, i.e. *ACWDDF*, is actually among the slowest.

It should be emphasized that some filters that appear in the 10 most efficient filters list but not in Table 13 still achieve a good compromise between effectiveness and efficiency. These include *MCWVMF*, *FMVMF*, *FFNRF*, *SVMF_rank*, *SVMF_mean*, and *FPGF*.

An examination of the distance measures (Minkowski, angular, directional-distance) with respect to effectiveness and efficiency shows that no distance measure completely outperforms the other two. However, it is interesting to note that among the 4 most effective filters, 2 are based on directional-distance (*ACWDDF*, *SDDF_rank*). Considering that only 8 of the 48 filters are based on directional-distance, the idea of combining the Minkowski and angular distance functions proves to be quite advantageous. On the other hand, as explained in Section 3.3, the filters based on the Minkowski distance are inherently more efficient than their angular and directional-distance counterparts. In fact, it can be seen from Table 12 that, except for the *FFNRF* and *ASBVDF_mean*, the most efficient 10 filters are all based on the Minkowski distance. In contrast, the most efficient angular filter (*ASBVDF_mean*) appears at the $7^{th}$ rank, whereas the most efficient directional-distance filter (*ASDDF_mean*) ranks $12^{th}$. This shows that if execution time is of prime importance, filters based on the Minkowski distance are the most obvious choice.

The unsatisfactory performance of the hybrid and adaptive fuzzy filters can be attributed to the fact that these filters introduce color artifacts by determining the output in a window as a linear or nonlinear combination of the input vectors. However, it should be noted that these filters are known to be more effective in the presence of Gaussian noise due to their averaging nature.

The reader should note that due to time constraints some filters in the literature were omitted from this study. Notable examples include the fast adaptive similarity based noise reduction filter (*FANRF*) [89], the fuzzy inference based vector filter (*FIVF*) [90], and the vector rank *M*-type *K*-nearest neighbor (*VRMKNNF*) [16]. The *FANRF* is based on the notion of similarity rather than distance. The similarity between two pixels can be calculated using various kernel functions which allows for more flexibility when designing filters tailored for particular applications. The *FIVF* employs a novel fuzzy inference system for noise detection and involves switching between the identity operation and the *L*-filter whose coefficients are determined using a fast constrained least-mean squares approach. The *VRMKNNF* is based on combined *RM*-estimators with different influence functions. It employs an adaptive non-parametric approach that determines the functional form of the probability density of the noise to improve the filtering performance.

### 4.2 Conclusions

This study presented a systematic survey of 48 impulsive noise removal filters using a unified notation. The filters were categorized into families and compared on a large image set in order to ensure an objective appraisal of their effectiveness and efficiency. A fast approximation for the inverse cosine function was introduced to allow for a more even comparison of efficiency. Furthermore, commonly used distance measures were compared and contrasted. Finally, recommendations for selecting filters that meet certain criteria were provided.

The implementations of the filters described in this article have been made publicly available as part of the Fourier image processing and analysis library, which can be downloaded from http://sourceforge.net/projects/fourier-ipal

## ACKNOWLEDGEMENTS

This work was supported by grants from NSF (#0216500-EIA), Texas Workforce Commission (#3204600182), and James A. Schlipmann Melanoma Cancer Foundation. Sources for the images in Figure 1 are as follows: http://pics.tech4learning.com (a-d, g),

EDRA Interactive Atlas of Dermoscopy (h), and Dr. Peter Alfeld (http://www.math.utah.edu/~pa/math/mandelbrot/bay.gif ) (i).

## REFERENCES

[1] Trussell H.J., Saber E., and Vrhel M.J. “Color Image Processing: Basics and Special Issue Overview,” IEEE Signal Processing Magazine 22(1), 14-22 (2005).

[2] Lukac R., Smolka B., Plataniotis K.N., and Venetsanopoulos A.N., “Selection Weighted Vector Directional Filters,” Computer Vision and Image Understanding 94(1/3), 140-167 (2004).

[3] Smolka B., Plataniotis K.N., and Venetsanopoulos A.N., “Nonlinear Techniques for Color Image Processing,” in *Nonlinear Signal and Image Processing: Theory, Methods, and Applications* (Editors: Barner K.E. and Arce G.R.), pp. 445-505, CRC Press, Boca Raton, FL (2004).

[4] Smolka B. and Chydzinski A., “Fast Detection and Impulsive Noise Removal in Color Images,” Real-Time Imaging 11(5/6), 389-402 (2005).

[5] Plataniotis K.N. and Venetsanopoulos A.N., *Color Image Processing and Applications*, Springer-Verlag, New York, NY (2000).

[6] Sharma G. and Trussell H.J., “Figures of Merit for Color Scanners,” IEEE Trans. on Image Processing 6(7), 990-1001 (1997).

[7] Lukac R., Smolka B., Martin K., Plataniotis K.N., and Venetsanopoulos A.N., “Vector Filtering for Color Imaging,” IEEE Signal Processing Magazine 22(1), 74-86 (2005).

[8] Astola J., Haavisto P., and Neuvo Y., “Vector Median Filters,” Proc. of the IEEE 78(4), 678-689 (1990).

[9] Barnett V. “The Ordering of Multivariate Data,” Journal of the Statistical Society of America A 139(3), 318-354 (1976).

[10] Pitas I. and Tsakalides P., “Multivariate Ordering in Color Image Filtering,” IEEE Trans. on Circuits and Systems for Video Technology 1(3), 247-259 (1991).

[11] Lukac R., Plataniotis K.N., Smolka B., and Venetsanopoulos A.N., “cDNA Microarray Image Processing Using Fuzzy Vector Filtering Framework,” Journal of Fuzzy Sets and Systems 152(1), 17-35 (2005).

[12] Lukac R., Plataniotis K.N., Smolka B., and Venetsanopoulos A.N., “A Multichannel Order-Statistic Technique for cDNA Microarray Image Processing,” IEEE Trans. on NanoBioscience 3(4), 272-285 (2004).

[13] Barni M., Bartolini F., and Cappellini V., “Image Processing for Virtual Restoration of Artworks,” IEEE Multimedia 7(2), 34-37 (2000).

[14] Lucchese L. and Mitra S.K., “A New Class of Chromatic Filters for Color Image Processing: Theory and Applications,” IEEE Trans. on Image Processing 13(4), 534-548 (2004).

[15] Lukac R., Fischer V., Motyl G., and Drutarovsky M., “Adaptive Video Filtering Framework,” International Journal of Imaging Systems and Technology 14(6), 223-237 (2004).

[16] Ponomaryov V.I., Gallegos-Funes F.J., and Rosales-Silva A., “Real-time Color Imaging Based on RM-filters for the Impulsive Noise Reduction,” Journal of Imaging Science and Technology 49(3), 205-219 (2005).

[17] Fischer V., Lukac R., and Martin K., "Cost-Effective Video Filtering Solution for Real-Time Vision Systems," EURASIP Journal on Applied Signal Processing 13, 2026-2042 (2005).

[18] Viero T., Oistamo K., and Neuvo Y., "Three-Dimensional Median-Related Filters for Color Image Sequence Filtering," IEEE Trans. on Circuits and Systems for Video Technology 4(2), 129-142 (1994).

[19] Lukac R. and Plataniotis K.N., "A Taxonomy of Color Image Filtering and Enhancement Solutions," in *Advances in Imaging & Electron Physics Vol. 140* (Editor: P.W. Hawkes), pp. 187-264, Academic Press, San Diego, CA (2006).

[20] Regazzoni C.S. and Teschioni A., "A New Approach to Vector Median Filtering Based on Space Filling Curves," IEEE Trans. on Image Processing 6(7), 1025-1037 (1997).

[21] Vardavoulia M. I., Andreadis I., and Tsalides Ph., "A New Vector Median Filter for Colour Image Processing," Pattern Recognition Letters 22(6/7), 675-689 (2001).

[22] Barni M., Cappellini V., and Mecocci A., "The Use of Different Metrics in Vector Median Filtering: Application to Fine Arts and Paintings," Proc. of EUSIPCO'92, 1485-1488 (1992).

[23] Trahanias P.E. and Venetsanopoulos A.N., "Vector Directional Filters: A New Class of Multichannel Image Processing Filters," IEEE Trans. on Image Processing 2(4), 528-534 (1993).

[24] Trahanias P.E, Karakos D., and Venetsanopoulos A.N., "Directional Processing of Color Images: Theory and Experimental Results," IEEE Trans. on Image Processing 5(6), 868-880 (1996).

[25] Astola J. and Kuosmanen P., "Fundamentals of Nonlinear Digital Filtering," CRC Press, Boca Raton, FL (1997).

[26] Karakos D.G. and Trahanias P.E., "Combining Vector Median and Vector Directional Filters: The Directional Distance Filters," Proc. of the IEEE ICIP Conf., 171-174 (1995).

[27] Karakos D.G. and Trahanias P.E., "Generalized Multichannel Image Filtering Structures," IEEE Trans. on Image Processing 6(7), 1038-1045 (1997).

[28] Plataniotis K.N., Androutsos D., and Venetsanopoulos A.N., "Content-Based Color Image Filters," Electronics Letters 33(3), 202-203 (1997).

[29] Ekman G., "A Direct Method for Multidimensional Ratio Scaling," Psychometrika 28, 33-41 (1963).

[30] Plataniotis K.N., Androutsos D., and Venetsanopoulos A.N., "Fuzzy Adaptive Filters for Multichannel Image Processing," Signal Processing 55(1), 93-106 (1996).

[31] Plataniotis K.N., Androutsos D., and Venetsanopoulos A.N., "Adaptive Fuzzy Systems for Multichannel Signal Processing," Proc. of the IEEE 87(9), 1601-1622 (1999).

[32] Chatzis V. and Pitas I., "Fuzzy Scalar and Vector Median Filters Based on Fuzzy Distances," IEEE Trans. on Image Processing 8(5), 731-734 (1999).

[33] Bezdek J.C. and Pal S.K., *Fuzzy Models for Pattern Recognition*, IEEE Press, New York, NY (1992).

[34] Plataniotis K.N., Androutsos D., Vinayagamoorthy S., and Venetsanopoulos A.N., "A Nearest Neighbor Multichannel Filter," Electronics Letters 31(22), 1910-1911 (1995).

[35] Plataniotis K.N., Androutsos D., Vinayagamoorthy S., and Venetsanopoulos A.N., "An Adaptive Nearest Neighbor Multichannel Filter," IEEE Trans. on Circuits and Systems for Video Technology 6(6), 699-703 (1996).

[36] Khriji L. and Gabbouj M., "Adaptive Fuzzy Order Statistics-Rational Hybrid Filters for Color Image Processing," Fuzzy Sets and Systems 128(1), 35-46 (2002).

[37] Ma Z., Wu H.R., and Qiu B., "A Robust Structure-Adaptive Hybrid Vector Filter Color Image Restoration," IEEE Trans. on Image Processing 14(12), 1990-2001 (2005).

[38] Gabbouj M. and Cheikh F.A., "Vector Median-Vector Directional Hybrid Filter for Color Image Restoration," Proc. of EUSIPCO'96 2, 879-882 (1996).

[39] Khriji L. and Gabbouj M., "A Class of Multichannel Image Processing Filters," Electronics Letters 35(4), 285-287 (1999).

[40] Khriji L. and Gabbouj M., "Vector Median-Rational Hybrid Filters for Multichannel Image Processing," IEEE Signal Processing Letters 6(7), 186-190 (1999).

[41] Khriji L. and Gabbouj M., "A New Class of Multichannel Image Processing Filters: Vector Median-Rational Hybrid Filters," IEICE Trans. on Information and Systems E82-D(12), 1589-1596 (1999).

[42] Khriji L. and Gabbouj M., "Multichannel Image Processing Using Fuzzy Vector Median-Rational Hybrid Filters," Proc. of EUSIPCO'00, 1345-1348 (2000).

[43] Khriji L. and Gabbouj M., "Rational-Based Adaptive Fuzzy Filters," Int. Journal of Computational Cognition 2(1), 113-132 (2004).

[44] Smolka B. and Plataniotis K.N., "Soft-Switching Adaptive Technique of Impulsive Noise Removal in Color Images," Proc. of the 2nd Int. Conf. on Image Analysis and Recognition (ICIAR 2005), Lecture Notes in Computer Science 3656, 686-693 (2005).

[45] Smolka B., Bieda R., Plataniotis K.N., and Lukac R., "Adaptive Soft-Switching Filter For Impulsive Noise Suppression in Color Images," Proc. of EUSIPCO'05 (2005).

[46] Smolka B., Plataniotis K.N., Lukac R., and Venetsanopoulos A.N., "New Class of Impulsive Noise Reduction Filters Based on Kernel Density Estimation," Proc. of the 28th IEEE Int. Conf. on Acoustics, Speech & Signal Processing (ICASSP'03) 3, 721-724 (2003).

[47] Smolka B., Lukac R., Plataniotis K.N., and Venetsanopoulos A.N., "Application of Kernel Density Estimation for Color Image Filtering," Proc. of Visual Communication and Image Processing (VCIP'03), SPIE Vol. 5150, 1650-1656 (2003).

[48] Smolka B., Plataniotis K.N., Lukac R., and Venetsanopoulos A.N., "Kernel Density Estimation Based Impulsive Noise Reduction Filter," Proc. of the IEEE Int. Conf. on Image Processing (ICIP'03) 2, 137-140 (2003).

[49] Oistamo K., Liu Q., Grundstrom M., Neuvo Y., "Weighted Vector Median Operation for Filtering Multispectral Data," Proc. of the IEEE Int. Conf. on Systems Engineering, 16-19 (1992).

[50] Lukac R., Plataniotis K.N., Smolka B., and Venetsanopoulos A.N., "Generalized Selection Weighted Vector Filters," EURASIP Journal of Applied Signal Processing 12, 1870-1885 (2004).

[51] Lukac R. Plataniotis K.N., and Venetsanopoulos A.N., "Color Image Denoising Using Evolutionary Computation," Int. Journal of Imaging Systems and Technology 15(5), 236-251 (2005).

[52] Lucat L., Siohan P., and Barba D., “Adaptive and Global Optimization Methods for Weighted Vector Median Filters,” Signal Processing: Image Communications 17(7), 509-524 (2002).

[53] Shen Y. and Barner K.E., “Fast Adaptive Optimization of Weighted Vector Median Filters,” IEEE Trans. on Signal Processing 54(7), 2497-2510 (2006).

[54] Lukac R., Plataniotis K.N., Smolka B., and Venetsanopoulos A.N., “Weighted Vector Median Optimization,” Proc. of the 4th EURASIP Conf. focused on Video/Image Processing and Multimedia Communications (EC-VIP-MC 2003) 1, 227-232 (2003).

[55] Lukac R., “Adaptive Color Image Filtering Based on Center-Weighted Vector Directional Filters,” Multidimensional Systems and Signal Processing 15(2), 169-196 (2004).

[56] Ma Z., Wu H.R., and Feng D., “Partition-Based Vector Filtering Technique for Suppression of Noise in Digital Color Images,” IEEE Trans. on Image Processing 15(8), 2324-2342 (2006).

[57] Lukac R., “Optimised Directional Distance Filter,” Machine Graphics and Vision 11(2/3), 311-326 (2002).

[58] Lukac R. and Marchevsky S., “Adaptive Vector LUM Smoother,” Proc. of the IEEE Int. Conf. on Image Processing (ICIP’01) 1, 878-881 (2001).

[59] Smolka B., “Efficient Modification of the Central Weighted Vector Median Filter,” Proc. of the 24th DAGM Symposium on Pattern Recognition, Lecture Notes in Computer Science 2449, 166-173 (2002).

[60] Smolka B., Lukac R., and Plataniotis K.N., "New Algorithm for Noise Attenuation in Color Images Based on the Central Weighted Vector Median Filter," Proc. of the 9th Int. Workshop on Systems, Signals and Image Processing (IWSSIP'02), 544-548 (2002).

[61] Lukac R., Smolka B., Plataniotis K.N., and Venetsanopoulos A.N., "Entropy Vector Median Filter," Proc. of the 1st Iberian Conf. on Pattern Recognition and Image Analysis (IbPRIA), Lecture Notes in Computer Science 2652, 1117-1125 (2003).

[62] Lukac R., Smolka B., Plataniotis K.N., and Venetsanopoulos A.N., "Generalized Entropy Vector Filters," Proc. of the 4th EURASIP EC-VIP-MC, Video, Image Processing and Multimedia Communications Conf., 239-244 (2003).

[63] Beghdadi A. and Khellaf A., "A Noise-Filtering Method Using a Local Information Measure," IEEE Trans. on Image Processing 6(6), 879-882 (1997).

[64] Lukac R., Smolka B., Plataniotis K.N., and Venetsanopoulos A.N., "Three-Dimensional Entropy Vector Median Filter for Color Video Filtering," Proc. of Visual Communication and Image Processing (VCIP'03), SPIE Vol. 5150, 1642-1649 (2003).

[65] Kenney C., Deng Y., Manjunath B.S., and Hewer G., "Peer Group Image Enhancement," IEEE Trans. on Image Processing 10(2), 326-334 (2001).

[66] Lukac R., Smolka B., Plataniotis K.N., Venetsanopoulos A.N., and Zavarsky P., "Angular Multichannel Sigma Filter," Proc. of the IEEE Int. Conf. on Acoustics, Speech, and Signal Processing (ICASSP'03) 3, 745-748 (2003).

[67] Lukac R., Smolka B., Plataniotis K.N., and Venetsanopoulos A.N., "A Variety of Multichannel Sigma Filters," Proc. of the SPIE 5146, 244-253 (2003).

[68] Lukac R., Smolka B., Plataniotis K.N., and Venetsanopoulos A.N., "Generalized Adaptive Vector Sigma Filters," Proc. of the Int. Conf. on Multimedia and Expo (ICME'03) 1, 537-540 (2003).

[69] Lukac R., Smolka B., Plataniotis K.N., and Venetsanopoulos A.N., "Vector Sigma Filters for Noise Detection and Removal in Color Images," Journal of Visual Communication and Image Representation 17(1), 1-26 (2006).

[70] Lukac R., Plataniotis K.N., Venetsanopoulos A.N., and Smolka B., "A Statistically-Switched Adaptive Vector Median Filter," Journal of Intelligent and Robotic Systems 42(4), 361-391 (2005).

[71] Lee J.S., "Digital Image Smoothing and the Sigma Filter," Computer Vision, Graphics and Image Processing 24(2), 255-269 (1983).

[72] Wilks S.S., "Certain Generalizations in the Analysis of Variance," Biometrika 24(3/4), 471-494 (1932).

[73] Moore M.S., Gabbouj M., and Mitra S.K., "Vector SD-ROM Filter for Removal of Impulse Noise from Color Images," Proc. of the 2nd EURASIP Conf. focused on DSP for Multimedia Communications and Services (ECMCS'99) (1999)

[74] Abreu E., Lightstone M., Mitra S.K., and Arakawa K., "A New Efficient Approach for the Removal of Impulse Noise from Highly Corrupted Images," IEEE Trans. on Image Processing 5(6), 1012-1025 (1996).

[75] Plataniotis K.N., Androutsos D., Vinayagamoorthy S., and Venetsanopoulos A.N., "Color Image Processing Using Adaptive Multichannel Filters," IEEE Trans. on Image Processing 6(7), 933-949 (1997).

[76] Plataniotis K.N., Androutsos D., and Venetsanopoulos A.N., "Adaptive Multichannel Filters for Colour Image Processing," Signal Processing: Image Communication 11(3), 171-177 (1998).

[77] Duda R.O., Hart P.E., and Stork D.G., *Pattern Classification* (Second Edition), Wiley-Interscience, New York, NY (2000).

[78] Smolka B., Szczepanski M., Plataniotis K.N., and Venetsanopoulos A.N., "Fast Modified Vector Median Filter," Proc. of the 9th Int. Conf. on Computer Analysis of Images and Patterns, Lecture Notes in Computer Science 2124, 570-580 (2001).

[79] Smolka B., Szczepanski M., Plataniotis K.N., and Venetsanopoulos A.N., "On the Fast Modification of the Vector Median Filter," Proc. of the 16th Int. Conf. on Pattern Recognition (ICPR'02) 3, 931-934 (2002).

[80] Lukac R., "Adaptive Vector Median Filtering," Pattern Recognition Letters 24(12), 1889-1899 (2003).

[81] Lukac R., "Color Image Filtering by Vector Directional Order-Statistics," Pattern Recognition and Image Analysis 12(3), 279-285 (2002).

[82] Morillas S., Gregori V., Peris-Fajarnes G., and Latorre P., "A New Vector Median Filter Based on Fuzzy Metrics," Proc. of the 2nd Int. Conf. on Image Analysis and Recognition (ICIAR'05), Lecture Notes in Computer Science 3656, 82-91 (2005).

[83] Morillas S., Gregori V., Peris-Fajarnes G., and Latorre P., "A Fast Impulsive Noise Color Image Filter Using Fuzzy Metrics," Real-Time Imaging 11(5/6), 417-428 (2005).

[84] George A. and Veeramani P., "On Some Results in Fuzzy Metric Spaces," Fuzzy Sets and Systems 64(3), 395-399 (1994).

[85] Cody W.J. and Waite W., *Software Manual for the Elementary Functions*, Prentice-Hall, Englewood Cliffs, NJ (1980).

[86] Muller J.-M., *Elementary Functions: Algorithms and Implementation* (Second Edition), Birkhäuser, Boston, MA (2006).

[87] Barni M., "A Fast Algorithm for 1-Norm Vector Median Filtering," IEEE Trans. on Image Processing 6(10), 1452-1455 (1997).

[88] Barni M., Buti F., Bartolini F., and Cappellini V., "A Quasi-Euclidean Norm to Speed up Vector Median Filtering," IEEE Trans. on Image Processing 9(10), 1704-1709 (2000).

[89] Smolka B., Lukac R., Chydzinski A., Plataniotis K.N., and Wojciechowski K., "Fast Adaptive Similarity Based Impulsive Noise Reduction Filter," Real-Time Imaging 9(4), 261-276 (2003).

[90] Hore E.S., Qiu B., and Wu H.R., "Improved Vector Filtering for Color Images Using Fuzzy Noise Detection," Optical Engineering 42(6), 1656-1664 (2003).

TABLE LEGEND

**Table 1. Notations used in the study**

| Notation | Meaning |
|---|---|
| $N$ | Number of pixels in an image |
| $W$ | Filtering window |
| $n$ | Number of pixels in $W$ |
| $x_i$ | $i^{th}$ pixel in $W$ |
| $x_i^k$ | $k^{th}$ component of $x_i$ ($k$ = 1: Red, $k$ = 2: Green, $k$ = 3: Blue) |
| $x_{(i)}$ | Pixel with the $i^{th}$ ranking according to a particular ordering scheme |
| $x_f$ | Output of a particular filter 'f' within $W$ |
| $C = (n+1)/2$ | Index of the center pixel in $W$ |
| $\Vert x_i \Vert = \left( x_i^1 \cdot x_i^1 + x_i^2 \cdot x_i^2 + x_i^3 \cdot x_i^3 \right)^{1/2}$ | Euclidean norm of $x_i$ |
| $\bar{x} = x_{AMF} = \frac{1}{n} \sum_{i=1}^{n} x_i$ | Mean vector within $W$. Also, the output of the Arithmetic Mean Filter ($AMF$). |
| $< x_i, x_j > = x_i^1 \cdot x_j^1 + x_i^2 \cdot x_j^2 + x_i^3 \cdot x_j^3$ | Inner product between $x_i$ and $x_j$ |
| $D(x_i, x_j)$ | Distance between $x_i$ and $x_j$ according to a particular measure |
| $L_p(x_i, x_j) = \Vert x_i - x_j \Vert_p = \left( \sum_{k=1}^{3} \left\vert x_i^k - x_j^k \right\vert^p \right)^{1/p}$ | Minkowski distance between $x_i$ and $x_j$ |
| $l(i) = l_p(i) = \sum_{j=1}^{n} L_p(x_i, x_j)$ | Cumulative Minkowski distance associated with $x_i$ |
| $A(x_i, x_j) = \cos^{-1}\left( \frac{< x_i, x_j >}{\Vert x_i \Vert \cdot \Vert x_j \Vert} \right)$ | Angular distance between $x_i$ and $x_j$ |
| $a(i) = \sum_{j=1}^{n} A(x_i, x_j)$ | Cumulative angular distance associated with $x_i$ |
| $d(i) = \left( \sum_{j=1}^{n} A(x_i, x_j) \right)^{\gamma} \cdot \left( \sum_{j=1}^{n} L_p(x_i, x_j) \right)^{1-\gamma}$ | Cumulative directional distance associated with $x_i$ |

**Table 2. Basic Vector Filters**

| Filter | Formulation |
|---|---|
| *VMF* | $x_{VMF} = \underset{x_i \in W}{argmin}(l(i))$ |
| *ATVMF* | $x_{ATVMF} = \frac{1}{1+\alpha}\sum_{i=1}^{1+\alpha} x_{(i)} \quad , \quad \alpha \in [0, n-1]$ |
| *BVDF* | $x_{BVDF} = \underset{x_i \in W}{argmin}(a(i))$ |
| *DDF* | $x_{DDF} = \underset{x_i \in W}{argmin}(d(i))$ |
| *CBRF* | $x_{CBRF} = \underset{x_i \in W}{argmin}\sum_{j=1}^{n} G(x_i, x_j)$ <br> $G(x_i, x_j) = \left( \frac{\lVert x_i \rVert^2 + \lVert x_j \rVert^2 - 2\lVert x_i \rVert \lVert x_j \rVert \cos(\theta)}{\lVert x_i \rVert^2 + \lVert x_j \rVert^2 + 2\lVert x_i \rVert \lVert x_j \rVert \cos(\theta)} \right)^{1/2}$ |

**Table 3. Hybrid Vector Filters**

<table>
<tr><th>Filter</th><th colspan="2">Formulation</th></tr>
<tr><td>EXVMF</td><td colspan="2">$x_{EXVMF} = \begin{cases} x_{AMF} & if\ l(x_{AMF}) \le l(x_{VMF}) \\ x_{VMF} & otherwise \end{cases}$</td></tr>
<tr><td>HDF</td><td colspan="2">$x_{HDF} = \begin{cases} x_{VMF} & if\ x_{VMF} = x_{BVDF} \\ \dfrac{\lVert x_{VMF} \rVert}{\lVert x_{BVDF} \rVert} \cdot x_{BVDF} & otherwise \end{cases}$</td></tr>
<tr><td>AHDF</td><td colspan="2">$x_{AHDF} = \begin{cases} x_{VMF} & if\ x_{VMF} = x_{BVDF} \\ x_{out1} & if\ l(x_{out1}) \le l(x_{out2}) \\ x_{out2} & otherwise \end{cases}$<br>$x_{out1} = \dfrac{\lVert x_{VMF} \rVert}{\lVert x_{BVDF} \rVert} \cdot x_{BVDF}$ , $x_{out2} = \dfrac{\lVert x_{AMF} \rVert}{\lVert x_{BVDF} \rVert} \cdot x_{BVDF}$</td></tr>
<tr><td>VMRHF</td><td>$x_{VMRHF} = x_{CWVMF} + \dfrac{\alpha_1 \cdot x_{VMF_1} + \alpha_2 \cdot x_{CWVMF} + \alpha_3 \cdot x_{VMF_2}}{\beta_1 + \beta_2 \cdot \lVert x_{VMF_1} - x_{VMF_2} \rVert}$</td><td rowspan="4">$\sum_{i=1}^{3} \alpha_i = 0$</td></tr>
<tr><td>FVMRHF</td><td>$x_{FVMRHF} = x_{FCWVMF} + \dfrac{\alpha_1 \cdot x_{FVMF_1} + \alpha_2 \cdot x_{FCWVMF} + \alpha_3 \cdot x_{FVMF_2}}{\beta_1 + \beta_2 \cdot \lVert x_{FVMF_1} - x_{FVMF_2} \rVert}$</td></tr>
<tr><td>FVDRHF</td><td>$x_{FVDRHF} = x_{FCWVDF} + \dfrac{\alpha_1 \cdot x_{FVDF_1} + \alpha_2 \cdot x_{FCWVDF} + \alpha_3 \cdot x_{FVDF_2}}{\beta_1 + \beta_2 \cdot A(x_{FVDF_1}, x_{FVDF_2})}$</td></tr>
<tr><td>FDDRHF</td><td>$x_{FDDRHF} = x_{FCWDDF} + \dfrac{\alpha_1 \cdot x_{FDDF_1} + \alpha_2 \cdot x_{FCWDDF} + \alpha_3 \cdot x_{FDDF_2}}{\beta_1 + \beta_2 \cdot \left[ A^{\gamma}(x_{FDDF_1}, x_{FDDF_2}) \cdot \lVert x_{FDDF_1} - x_{FDDF_2} \rVert^{1-\gamma} \right]}$</td></tr>
<tr><td>KVMF</td><td colspan="2">$x_{KVMF} = \mu\left(\lVert x_C - x_{VMF} \rVert\right) \cdot x_C + \left(1 - \mu\left(\lVert x_C - x_{VMF} \rVert\right)\right) \cdot x_{VMF}$<br>$\mu(d) = \exp\left(-d/h\right)$ , $h \cong \dfrac{\beta}{\left(\sum_{i=1}^{N} \lVert x_i - \overline{x} \rVert^2 / 8N^2\right)^{1/2}}$</td></tr>
</table>

**Table 4. Adaptive Center-Weighted Vector Filters**

| Filter | Formulation |
|---|---|
| *MCWVMF* | $x_{MCWVMF} = \begin{cases} x_{VMF} & if\ l(x_{VMF}) < w \cdot l(C) \\ x_C & otherwise \end{cases} \quad , \quad w \in [0,1]$ |
| *ACWVMF* | $x_{ACWVMF} = \begin{cases} x_{VMF} & if\ \sum_{k=\lambda}^{\lambda+2} \left\| x_{CWVMF^k} - x_C \right\| > T \\ x_C & otherwise \end{cases} \quad , \quad \lambda \in [1, C-1]$ |
| *ACWVDF* | $x_{ACWVDF} = \begin{cases} x_{BVDF} & if\ \sum_{k=\lambda}^{\lambda+2} A(x_{CWVDF^k}, x_C) > T \\ x_C & otherwise \end{cases} \quad , \quad \lambda \in [1, C-1]$ |
| *ACWDDF* | $x_{ACWDDF} = \begin{cases} x_{DDF} & if\ \sum_{k=\lambda}^{\lambda+2} A^{\gamma}(x_{CWDDF^k}, x_C) \cdot \left\| x_{CWDDF^k} - x_C \right\|^{1-\gamma} > T \\ x_C & otherwise \end{cases} \quad , \quad \lambda \in [1, C-1]$ |

**Table 5. Entropy Vector Filters**

| Filter | Formulation |
|---|---|
| *EVMF* | $x_{EVMF} = \begin{cases} x_{VMF} & if\ P_C > T_C \\ x_C & otherwise \end{cases}$ <br> $P_i = \frac{\lVert x_i - \overline{x} \rVert}{\sum_{j=1}^{n} \lVert x_j - \overline{x} \rVert}$ , $T_i = \frac{-P_i\, logP_i}{-\sum_{j=1}^{n} P_j\, logP_j}$ |
| *EBVDF* | $x_{EBVDF} = \begin{cases} x_{BVDF} & if\ P_C > T_C \\ x_C & otherwise \end{cases}$ <br> $P_i = \frac{A(x_i, \overline{x})}{\sum_{j=1}^{n} A(x_j, \overline{x})}$ , $T_i = \frac{-P_i\, logP_i}{-\sum_{j=1}^{n} P_j\, logP_j}$ |
| *EDDF* | $x_{EDDF} = \begin{cases} x_{DDF} & if\ P_C > T_C \\ x_C & otherwise \end{cases}$ <br> $P_i = \frac{A(x_i, \overline{x})^{\gamma} \lVert x_i - \overline{x} \rVert^{1-\gamma}}{\sum_{j=1}^{n} A(x_j, \overline{x})^{\gamma} \lVert x_j - \overline{x} \rVert^{1-\gamma}}$ , $T_i = \frac{-P_i\, logP_i}{-\sum_{j=1}^{n} P_j\, logP_j}$ |

**Table 6. Peer Group Vector Filters**

| Filter | Formulation |
|---|---|
| *PGF* | $c(i) = \lVert x_C - x_i \rVert \quad for\ \ i = 1, 2, \ldots, n$ <br> $\delta(i) = c_{(i+1)} - c_{(i)} \quad for\ \ i = 1, 2, \ldots, m = (\sqrt{n} + 1)/2$ <br> $x_{PGF} = \begin{cases} x_{VMF} & if\ \exists i \in [1, m]\ \ s.t.\ \ \delta(i) > T \\ x_C & otherwise \end{cases}$ |
| *FPGF* | $x_{FPGF} = \begin{cases} x_{VMF} & if\ \ \left\lvert \left\{ x_{i \neq C} \in W\ \ s.t.\ \lVert x_C - x_i \rVert \leq T \right\} \right\rvert < m \\ x_C & otherwise \end{cases}$ |

**Table 7. Vector Sigma Filters**

| Filter | Formulation |
|---|---|
| *SVMF_mean* | $x_{SVMF_mean} = \begin{cases} x_{VMF} & if\ l(C) \geq (1+\lambda/n)\cdot l(\overline{x}) \\ x_C & otherwise \end{cases}$ |
| *SVMF_rank* | $x_{SVMF_rank} = \begin{cases} x_{VMF} & if\ l(C) \geq \left(1+\lambda/(n-1)\right)\cdot l(x_{VMF}) \\ x_C & otherwise \end{cases}$ |
| *SBVDF_mean* | $x_{SBVDF_mean} = \begin{cases} x_{BVDF} & if\ a(C) \geq (1+\lambda/n)\cdot a(\overline{x}) \\ x_C & otherwise \end{cases}$ |
| *SBVDF_rank* | $x_{SBVDF_rank} = \begin{cases} x_{BVDF} & if\ a(C) \geq \left(1+\lambda/(n-1)\right)\cdot a(x_{BVDF}) \\ x_C & otherwise \end{cases}$ |
| *SDDF_mean* | $x_{SDDF_mean} = \begin{cases} x_{DDF} & if\ d(C) \geq (1+\lambda/n)\cdot d(\overline{x}) \\ x_C & otherwise \end{cases}$ |
| *SDDF_rank* | $x_{SDDF_rank} = \begin{cases} x_{DDF} & if\ d(C) \geq \left(1+\lambda/(n-1)\right)\cdot d(x_{DDF}) \\ x_C & otherwise \end{cases}$ |
| *ASVMF_mean* | $x_{ASVMF_mean} = \begin{cases} x_{VMF} & if\ \left\Vert x_C - \overline{x} \right\Vert \geq \sigma \\ x_C & otherwise \end{cases}$ <br> $\sigma^2 = \frac{1}{n}\sum_{i=1}^{n}\left\Vert x_i - \overline{x} \right\Vert^2$ |
| *ASVMF_rank* | $x_{ASVMF_rank} = \begin{cases} x_{VMF} & if\ \left\Vert x_C - x_{VMF} \right\Vert \geq \sigma \\ x_C & otherwise \end{cases}$ <br> $\sigma^2 = \frac{1}{n-1}\sum_{i=1}^{n}\left\Vert x_i - x_{VMF} \right\Vert^2$ |
| *ASBVDF_mean* | $x_{ASBVDF_mean} = \begin{cases} x_{BVDF} & if\ A(x_C, \overline{x}) \geq \sigma \\ x_C & otherwise \end{cases}$ <br> $\sigma^2 = \frac{1}{n}\sum_{i=1}^{n} A^2(x_i, \overline{x})$ |
| *ASBVDF_rank* | $x_{ASBVDF_rank} = \begin{cases} x_{BVDF} & if\ A(x_C, x_{BVDF}) \geq \sigma \\ x_C & otherwise \end{cases}$ <br> $\sigma^2 = \frac{1}{n-1}\sum_{i=1}^{n} A^2(x_i, x_{BVDF})$ |
| *ASDDF_mean* | $x_{ASDDF_mean} = \begin{cases} x_{DDF} & if\ A^{\gamma}(x_C, \overline{x})\cdot \left\Vert x_C - \overline{x} \right\Vert^{1-\gamma} \geq \sigma \\ x_C & otherwise \end{cases}$ <br> $\sigma^2 = \left(\frac{1}{n}\sum_{i=1}^{n} A^2(x_i, \overline{x})\right)^{\gamma} \left(\frac{1}{n}\sum_{i=1}^{n}\left\Vert x_i - \overline{x} \right\Vert^2\right)^{1-\gamma}$ |
| *ASDDF_rank* | $x_{ASDDF_rank} = \begin{cases} x_{DDF} & if\ \ A^{\gamma}(x_C, x_{DDF}) \lVert x_C - x_{DDF} \rVert^{1-\gamma} \geq \sigma \\ x_C & otherwise \end{cases}$ <br> $\sigma^2 = \left( \frac{1}{n-1} \sum_{i=1}^{n} A^2(x_i, x_{DDF}) \right)^{\gamma} \left( \frac{1}{n-1} \sum_{i=1}^{n} \lVert x_i - x_{DDF} \rVert^2 \right)^{1-\gamma}$ |

**Table 8. Miscellaneous Vector Filters**

| Filter | Formulation |
|---|---|
| *VSDROMF* | $x_{VSDROMF} = \begin{cases} x_{VMF} & if\ \exists i \in \{1,2,3,4\}\ \ s.t.\ \ \left\Vert x_C - x_{(i)} \right\Vert > T_i \\ x_C & otherwise \end{cases}$ <br> $T_1 \le T_2 \le T_3 \le T_4$ |
| *AMNF* | $x_{AMNF} = \sum_{i=1}^{n} x_i \left( \dfrac{h_i^{-c} K\left(\dfrac{x_C - x_i}{h_i}\right)}{\sum_{j=1}^{n} h_j^{-c} K\left(\dfrac{x_C - x_j}{h_j}\right)} \right)$ <br> $h_i = n^{-k/c} \sum_{j=1}^{n} \left\Vert x_i - x_j \right\Vert_1$ |
| *FMVMF* | $x_{FMVMF} = \begin{cases} x_{k^*} & if\ \left( \sum_{i=1}^{n} \left\Vert x_C - x_i \right\Vert - \sum_{\substack{i=1 \\ i \ne C}}^{n} \left\Vert x_{k^*} - x_i \right\Vert \right) > T \\ x_C & otherwise \end{cases}$ <br> $x_{k^*} = \underset{x_k \in W}{argmin} \sum_{\substack{i=1 \\ i \ne C}}^{n} \left\Vert x_k - x_i \right\Vert$ |
| *AVMF* | $x_{AVMF} = \begin{cases} x_{VMF} & if\ \left\Vert x_C - \dfrac{1}{k} \sum_{i=1}^{k} x_{(i)} \right\Vert > T \\ x_C & otherwise \end{cases}$ |
| *ABVDF* | $x_{ABVDF} = \begin{cases} x_{BVDF} & if\ \ A\left( x_C\, , \dfrac{1}{k} \sum_{i=1}^{k} x_{(i)} \right) > T \\ x_C & otherwise \end{cases}$ |
| *FFNRF* | $x_{FFNRF} = \begin{cases} x_{k^*} & if\ \ \sum_{i=1}^{n} M(x_C, x_i) < \sum_{\substack{i=1 \\ i \ne C}}^{n} M(x_{k^*}, x_i) \\ x_C & otherwise \end{cases}$ <br> $x_{k^*} = \underset{x_k \in W}{argmax} \sum_{\substack{i=1 \\ i \ne C}}^{n} M(x_k, x_i) \quad , \quad M^{\alpha}(x_i, x_j) = \prod_{k=1}^{3} \left( \dfrac{min(x_i^{\,k}, x_j^{\,k}) + K}{max(x_i^{\,k}, x_j^{\,k}) + K} \right)^{\alpha}$ |

**Table 9. Comparison of the filters based on the *MAE* measure (AR: average ranking)**

| MAE | Uncorrelated Noise | | | | | | Correlated Noise | | | | | |
|---|---|---|---|---|---|---|---|---|---|---|---|---|
| | 5% | | 10% | | 15% | | 5% | | 10% | | 15% | |
| Rank | Filter | AR | Filter | AR | Filter | AR | Filter | AR | Filter | AR | Filter | AR |
| 0 | acwddf | 1.56 | acwddf | 1.71 | acwddf | 1.78 | acwddf | 1.75 | acwddf | 1.84 | acwddf | 2.35 |
| 1 | pgf | 2.29 | pgf | 3.31 | sddf_rank | 3.27 | pgf | 1.96 | pgf | 2.52 | pgf | 3.32 |
| 2 | mcwvmf | 3.31 | mcwvmf | 5.07 | pgf | 5.16 | mcwvmf | 3.30 | acwvmf | 4.58 | sddf_rank | 3.95 |
| 3 | acwvdf | 5.03 | sddf_rank | 5.08 | acwvmf | 5.34 | acwvmf | 4.75 | sddf_rank | 5.23 | acwvmf | 3.98 |
| 4 | acwvmf | 5.06 | acwvmf | 5.30 | sddf_mean | 6.82 | acwvdf | 5.89 | mcwvmf | 5.25 | svmf_rank | 6.98 |
| 5 | abvdf | 7.07 | acwvdf | 6.21 | acwvdf | 7.82 | avmf | 6.75 | sddf_mean | 8.07 | sddf_mean | 7.85 |
| 6 | sddf_rank | 7.66 | sddf_mean | 7.74 | svmf_rank | 8.75 | sddf_rank | 7.68 | acwvdf | 8.08 | ffnrf | 8.75 |
| 7 | avmf | 7.85 | abvdf | 9.84 | asddf_rank | 10.34 | ffnrf | 8.13 | ffnrf | 8.88 | svmf_mean | 9.51 |
| 8 | ffnrf | 8.70 | ffnrf | 10.32 | ffnrf | 10.63 | abvdf | 8.61 | svmf_rank | 9.66 | acwvdf | 10.30 |
| 9 | sddf_mean | 9.50 | svmf_rank | 11.20 | mcwvmf | 11.04 | sddf_mean | 9.63 | avmf | 9.98 | fmvmf | 11.23 |
| 10 | sbvdf_rank | 11.65 | sbvdf_rank | 11.51 | svmf_mean | 11.49 | fpgf | 11.72 | svmf_mean | 11.52 | mcwvmf | 11.48 |
| 11 | fpgf | 12.12 | asddf_rank | 11.97 | sbvdf_rank | 11.83 | sbvdf_rank | 12.28 | abvdf | 12.39 | avmf | 12.13 |
| 12 | fmvmf | 12.91 | avmf | 12.81 | abvdf | 12.23 | svmf_rank | 12.39 | fmvmf | 12.40 | asddf_rank | 12.70 |
| 13 | svmf_rank | 13.31 | fmvmf | 13.15 | fmvmf | 12.47 | fmvmf | 12.51 | asddf_rank | 12.85 | fpgf | 13.98 |
| 14 | asddf_rank | 13.69 | svmf_mean | 13.17 | asddf_mean | 13.15 | asddf_rank | 13.75 | sbvdf_rank | 13.36 | sbvdf_rank | 15.11 |
| 15 | sbvdf_mean | 14.29 | asddf_mean | 14.20 | eddf | 14.47 | svmf_mean | 13.89 | fpgf | 13.55 | evmf | 15.14 |
| 16 | svmf_mean | 14.81 | fpgf | 14.57 | fpgf | 15.77 | sbvdf_mean | 15.81 | asddf_mean | 16.16 | abvdf | 15.20 |
| 17 | asddf_mean | 15.83 | sbvdf_mean | 16.32 | avmf | 16.49 | asddf_mean | 16.37 | eddf | 16.85 | asvmf_mean | 15.43 |
| 18 | eddf | 18.34 | eddf | 16.34 | evmf | 16.93 | eddf | 18.41 | asvmf_rank | 17.25 | asvmf_rank | 15.58 |
| 19 | asbvdf_rank | 18.89 | asvmf_rank | 18.62 | asvmf_mean | 17.50 | asvmf_rank | 18.95 | evmf | 17.58 | eddf | 15.88 |
| 20 | asvmf_rank | 19.90 | evmf | 18.79 | asvmf_rank | 17.63 | asbvdf_rank | 19.60 | asvmf_mean | 18.73 | asddf_mean | 16.71 |
| 21 | vsdromf | 20.43 | asbvdf_rank | 18.80 | sbvdf_mean | 18.75 | evmf | 19.94 | sbvdf_mean | 19.47 | vsdromf | 19.05 |
| 22 | evmf | 20.86 | asvmf_mean | 20.18 | asbvdf_rank | 19.22 | vsdromf | 20.22 | vsdromf | 20.20 | asbvdf_rank | 21.94 |
| 23 | ebvdf | 21.62 | vsdromf | 21.16 | vsdromf | 20.64 | asvmf_mean | 21.43 | asbvdf_rank | 20.62 | vmrhf | 22.27 |
| 24 | asvmf_mean | 22.38 | ebvdf | 23.31 | vmrhf | 23.51 | ebvdf | 23.40 | vmrhf | 23.85 | sbvdf_mean | 22.30 |
| 25 | asbvdf_mean | 22.70 | asbvdf_mean | 23.35 | asbvdf_mean | 24.03 | asbvdf_mean | 24.26 | asbvdf_mean | 25.05 | fvmrhf | 24.67 |
| 26 | vmrhf | 25.30 | vmrhf | 24.52 | ebvdf | 24.84 | vmrhf | 24.98 | ebvdf | 25.50 | kvmf | 24.80 |
| 27 | kvmf | 25.59 | kvmf | 26.29 | fvmrhf | 25.64 | kvmf | 25.52 | kvmf | 25.73 | asbvdf_mean | 26.73 |
| 28 | fvmrhf | 26.88 | fvmrhf | 26.43 | kvmf | 25.78 | fvmrhf | 26.64 | fvmrhf | 25.94 | fddrhf | 27.60 |
| 29 | fddrhf | 28.77 | fddrhf | 28.49 | fddrhf | 28.13 | fddrhf | 28.64 | fddrhf | 28.26 | ebvdf | 27.89 |
| 30 | vmf | 30.97 | vmf | 30.79 | vmf | 30.57 | vmf | 30.82 | vmf | 30.51 | vmf | 30.13 |
| 31 | cbrf | 32.31 | ddf | 32.60 | ddf | 32.31 | cbrf | 32.29 | exvmf | 32.27 | ddf | 32.01 |
| 32 | ddf | 32.67 | exvmf | 32.74 | exvmf | 32.54 | exvmf | 32.58 | ddf | 32.55 | exvmf | 32.12 |
| 33 | exvmf | 32.85 | cbrf | 32.97 | fovmf | 32.81 | ddf | 32.76 | cbrf | 33.07 | fovmf | 32.28 |
| 34 | amnfe | 34.02 | fovmf | 33.80 | fvmf | 32.91 | fovmf | 34.37 | fovmf | 33.46 | fvmf | 32.28 |
| 35 | fovmf | 34.55 | amnfe | 34.41 | cbrf | 33.79 | amnfe | 34.60 | fvmf | 33.90 | cbrf | 34.03 |
| 36 | ahdf | 35.06 | fvmf | 34.41 | amnfe | 35.30 | ahdf | 34.91 | ahdf | 35.01 | ahdf | 34.82 |
| 37 | fvmf | 35.57 | ahdf | 35.27 | ahdf | 35.45 | fvmf | 35.40 | amnfe | 35.52 | hdf | 35.74 |
| 38 | hdf | 36.19 | hdf | 36.34 | hdf | 36.39 | hdf | 36.12 | hdf | 36.06 | amnfe | 36.64 |
| 39 | amnfg | 38.56 | amnfg | 38.51 | amnfg | 38.63 | amnfg | 38.57 | amnfg | 38.67 | atvmf | 38.40 |
| 40 | atvmf | 39.83 | atvmf | 39.58 | atvmf | 39.28 | atvmf | 39.73 | atvmf | 39.22 | amnfg | 39.04 |
| 41 | annmf | 41.10 | annmf | 41.70 | fovdf | 41.56 | annmf | 41.12 | annmf | 41.79 | fovdf | 41.46 |
| 42 | fovdf | 43.15 | fovdf | 42.27 | gvdf | 42.07 | fovdf | 42.80 | fovdf | 42.15 | gvdf | 41.64 |
| 43 | annf | 43.86 | gvdf | 42.82 | annmf | 42.67 | gvdf | 43.49 | gvdf | 42.27 | annmf | 42.61 |
| 44 | fvdrhf | 43.86 | bvdf | 44.72 | bvdf | 44.39 | annf | 44.14 | bvdf | 44.30 | bvdf | 43.66 |
| 45 | gvdf | 44.15 | annf | 44.83 | fvdf | 44.71 | fvdrhf | 44.34 | fvdrhf | 45.26 | fvdf | 44.67 |
| 46 | bvdf | 44.87 | fvdrhf | 44.84 | fvdrhf | 45.38 | bvdf | 44.66 | annf | 45.32 | fvdrhf | 45.77 |
| 47 | fvdf | 46.13 | fvdf | 45.64 | annf | 45.79 | fvdf | 46.14 | fvdf | 45.32 | annf | 45.89 |

**Table 10. Comparison of the filters based on the *MSE* measure (AR: average ranking)**

| MSE | Uncorrelated Noise | | | | | | Correlated Noise | | | | | |
|---|---|---|---|---|---|---|---|---|---|---|---|---|
| | 5% | | 10% | | 15% | | 5% | | 10% | | 15% | |
| **Rank** | **Filter** | **AR** | **Filter** | **AR** | **Filter** | **AR** | **Filter** | **AR** | **Filter** | **AR** | **Filter** | **AR** |
| 0 | acwddf | 3.26 | acwddf | 3.96 | acwddf | 4.29 | pgf | 2.56 | pgf | 4.13 | acwvmf | 4.32 |
| 1 | pgf | 3.82 | sddf_rank | 4.37 | sddf_rank | 5.00 | acwddf | 3.88 | sddf_rank | 4.51 | fvmrhf | 5.76 |
| 2 | sddf_rank | 5.50 | acwvmf | 6.86 | acwvmf | 6.72 | sddf_rank | 5.38 | acwvmf | 4.68 | sddf_rank | 5.88 |
| 3 | acwvmf | 6.64 | pgf | 7.43 | fvmrhf | 6.94 | acwvmf | 5.48 | acwddf | 4.89 | acwddf | 5.99 |
| 4 | acwvdf | 7.37 | sddf_mean | 9.15 | vmrhf | 9.04 | mcwvmf | 7.12 | svmf_rank | 6.93 | pgf | 6.49 |
| 5 | mcwvmf | 7.76 | acwvdf | 9.47 | fddrhf | 9.07 | sddf_mean | 8.45 | fvmrhf | 8.98 | fddrhf | 7.73 |
| 6 | sddf_mean | 7.95 | svmf_rank | 9.95 | svmf_rank | 11.27 | svmf_rank | 9.04 | vmrhf | 10.65 | vmrhf | 7.85 |
| 7 | svmf_rank | 11.22 | fvmrhf | 10.50 | acwvdf | 12.43 | svmf_mean | 11.03 | svmf_mean | 10.77 | svmf_rank | 9.05 |
| 8 | abvdf | 11.35 | vmrhf | 12.37 | pgf | 12.99 | acwvdf | 12.30 | fddrhf | 11.20 | fpgf | 10.65 |
| 9 | asddf_rank | 12.91 | asddf_rank | 12.87 | fmvmf | 13.17 | fvmrhf | 12.78 | sddf_mean | 11.41 | fmvmf | 10.93 |
| 10 | svmf_mean | 13.07 | fddrhf | 13.33 | fpgf | 13.85 | ffnrf | 13.37 | fpgf | 12.90 | ffnrf | 12.50 |
| 11 | sbvdf_rank | 13.68 | svmf_mean | 13.35 | sddf_mean | 14.24 | vmrhf | 13.68 | ffnrf | 13.02 | kvmf | 14.35 |
| 12 | fvmrhf | 14.35 | sbvdf_rank | 14.48 | asddf_rank | 15.60 | fpgf | 13.88 | fmvmf | 13.92 | svmf_mean | 15.46 |
| 13 | vmrhf | 15.31 | abvdf | 15.13 | kvmf | 16.34 | asddf_rank | 15.01 | asvmf_mean | 16.77 | vsdromf | 16.11 |
| 14 | fpgf | 15.33 | fpgf | 15.51 | ffnrf | 16.68 | fddrhf | 15.50 | evmf | 16.89 | asvmf_mean | 18.10 |
| 15 | ffnrf | 15.59 | fmvmf | 16.02 | abvdf | 16.85 | avmf | 16.11 | kvmf | 17.73 | sddf_mean | 18.55 |
| 16 | asddf_mean | 15.71 | eddf | 16.64 | svmf_mean | 17.12 | fmvmf | 16.63 | acwvdf | 18.41 | fvmf | 18.64 |
| 17 | eddf | 16.22 | ffnrf | 16.69 | asvmf_mean | 17.54 | eddf | 17.47 | asvmf_rank | 19.06 | amnfe | 19.24 |
| 18 | fddrhf | 17.37 | asddf_mean | 17.63 | sbvdf_rank | 18.60 | sbvdf_rank | 17.74 | asddf_rank | 19.10 | evmf | 19.61 |
| 19 | fmvmf | 18.11 | evmf | 18.36 | evmf | 18.95 | evmf | 18.12 | vsdromf | 19.46 | fovmf | 20.67 |
| 20 | evmf | 20.03 | asvmf_mean | 18.80 | vsdromf | 18.99 | asvmf_mean | 19.40 | eddf | 20.93 | asvmf_rank | 20.92 |
| 21 | avmf | 20.40 | kvmf | 19.54 | eddf | 20.20 | asddf_mean | 19.63 | avmf | 21.33 | amnfg | 22.74 |
| 22 | kvmf | 21.30 | asvmf_rank | 21.11 | amnfe | 21.13 | abvdf | 19.70 | mcwvmf | 22.25 | avmf | 23.09 |
| 23 | asvmf_mean | 21.41 | mcwvmf | 21.98 | asddf_mean | 22.21 | asvmf_rank | 19.96 | amnfe | 22.33 | acwvdf | 23.25 |
| 24 | sbvdf_mean | 21.47 | vsdromf | 22.15 | fvmf | 22.30 | kvmf | 20.09 | sbvdf_rank | 23.37 | asddf_rank | 24.59 |
| 25 | asvmf_rank | 21.78 | amnfe | 24.66 | asvmf_rank | 22.31 | vsdromf | 22.30 | fvmf | 23.53 | exvmf | 25.09 |
| 26 | asbvdf_rank | 23.01 | asbvdf_rank | 26.11 | fovmf | 24.26 | amnfe | 25.94 | fovmf | 25.31 | ahdf | 25.57 |
| 27 | vsdromf | 24.10 | fvmf | 26.51 | amnfg | 24.36 | fvmf | 27.85 | amnfg | 25.65 | atvmf | 26.14 |
| 28 | amnfe | 27.17 | amnfg | 27.24 | exvmf | 28.33 | sbvdf_mean | 28.30 | asddf_mean | 26.40 | vmf | 26.16 |
| 29 | fvmf | 29.35 | avmf | 27.84 | ahdf | 28.69 | asbvdf_rank | 28.35 | abvdf | 26.57 | eddf | 26.71 |
| 30 | amnfg | 29.58 | fovmf | 28.11 | vmf | 29.76 | amnfg | 28.47 | exvmf | 29.07 | hdf | 28.41 |
| 31 | asbvdf_mean | 30.87 | sbvdf_mean | 28.27 | atvmf | 30.11 | fovmf | 29.51 | ahdf | 29.72 | ddf | 28.94 |
| 32 | ebvdf | 30.94 | exvmf | 31.37 | avmf | 30.48 | exvmf | 32.38 | atvmf | 30.82 | abvdf | 29.19 |
| 33 | fovmf | 31.06 | ahdf | 32.18 | asbvdf_rank | 30.60 | ahdf | 33.13 | vmf | 30.87 | sbvdf_rank | 30.32 |
| 34 | exvmf | 33.86 | vmf | 33.33 | hdf | 31.66 | vmf | 34.71 | hdf | 32.58 | cbrf | 30.78 |
| 35 | ahdf | 34.66 | atvmf | 33.70 | ddf | 32.95 | annmf | 35.21 | ddf | 33.67 | asddf_mean | 32.34 |
| 36 | fvdrhf | 35.51 | hdf | 34.99 | cbrf | 34.24 | atvmf | 35.44 | cbrf | 33.85 | fovdf | 32.39 |
| 37 | vmf | 36.37 | fvdrhf | 35.58 | sbvdf_mean | 35.10 | fvdrhf | 35.57 | annmf | 34.81 | annmf | 33.85 |
| 38 | annmf | 36.42 | cbrf | 36.23 | fvdrhf | 35.29 | hdf | 36.02 | asbvdf_rank | 34.97 | fvdf | 33.95 |
| 39 | atvmf | 37.11 | ddf | 36.66 | fovdf | 35.37 | cbrf | 36.59 | fvdrhf | 35.39 | gvdf | 34.31 |
| 40 | hdf | 37.44 | asbvdf_mean | 36.90 | mcwvmf | 35.71 | asbvdf_mean | 37.93 | sbvdf_mean | 36.82 | fvdrhf | 34.32 |
| 41 | cbrf | 37.88 | annmf | 37.15 | fvdf | 36.06 | ebvdf | 37.93 | fovdf | 37.21 | mcwvmf | 36.16 |
| 42 | ddf | 39.55 | ebvdf | 37.94 | annmf | 37.47 | ddf | 38.07 | fvdf | 37.54 | annf | 38.72 |
| 43 | annf | 40.23 | fvdf | 39.16 | gvdf | 38.03 | annf | 39.84 | gvdf | 38.56 | asbvdf_rank | 39.45 |
| 44 | fvdf | 41.66 | fovdf | 39.50 | annf | 39.67 | fvdf | 40.85 | annf | 39.76 | sbvdf_mean | 41.49 |
| 45 | fovdf | 42.30 | annf | 40.26 | asbvdf_mean | 40.30 | fovdf | 41.27 | asbvdf_mean | 41.92 | bvdf | 41.70 |
| 46 | gvdf | 43.61 | gvdf | 41.15 | ebvdf | 41.85 | gvdf | 42.17 | ebvdf | 43.13 | asbvdf_mean | 44.11 |
| 47 | bvdf | 46.41 | bvdf | 45.51 | bvdf | 43.88 | bvdf | 45.86 | bvdf | 44.23 | ebvdf | 45.43 |

**Table 11. Comparison of the filters based on the *NCD* measure (AR: average ranking)**

| NCD | Uncorrelated Noise | | | | | | Correlated Noise | | | | | |
|---|---|---|---|---|---|---|---|---|---|---|---|---|
| | 5% | | 10% | | 15% | | 5% | | 10% | | 15% | |
| Rank | Filter | AR | Filter | AR | Filter | AR | Filter | AR | Filter | AR | Filter | AR |
| 0 | acwddf | 1.79 | acwddf | 2.31 | acwddf | 2.54 | acwddf | 1.80 | acwddf | 2.22 | acwddf | 2.34 |
| 1 | pgf | 4.18 | sddf_rank | 3.29 | sddf_rank | 3.04 | pgf | 3.28 | sddf_rank | 3.62 | sddf_rank | 3.46 |
| 2 | acwvdf | 4.41 | acwvdf | 5.05 | acwvdf | 5.28 | acwvmf | 4.25 | acwvmf | 4.37 | acwvmf | 4.43 |
| 3 | acwvmf | 4.60 | acwvmf | 5.63 | acwvmf | 6.16 | acwvdf | 5.03 | pgf | 5.10 | acwvdf | 6.07 |
| 4 | mcwvmf | 5.35 | sddf_mean | 7.19 | fmvmf | 7.29 | mcwvmf | 5.33 | acwvdf | 5.63 | pgf | 6.32 |
| 5 | sddf_rank | 5.69 | pgf | 7.44 | svmf_rank | 7.78 | sddf_rank | 5.94 | svmf_rank | 7.82 | fmvmf | 6.58 |
| 6 | sddf_mean | 8.39 | sbvdf_rank | 8.10 | sbvdf_rank | 7.83 | ffnrf | 8.87 | sddf_mean | 8.05 | svmf_rank | 7.33 |
| 7 | abvdf | 8.48 | svmf_rank | 8.65 | sddf_mean | 8.54 | abvdf | 8.95 | fmvmf | 8.90 | sbvdf_rank | 9.52 |
| 8 | ffnrf | 10.09 | fmvmf | 9.44 | pgf | 10.02 | sddf_mean | 8.96 | sbvdf_rank | 9.35 | sddf_mean | 10.09 |
| 9 | sbvdf_rank | 10.33 | abvdf | 10.05 | abvdf | 10.30 | fmvmf | 10.33 | abvdf | 10.66 | abvdf | 10.69 |
| 10 | fmvmf | 10.44 | svmf_mean | 11.60 | svmf_mean | 12.03 | svmf_rank | 10.70 | ffnrf | 10.73 | ffnrf | 10.76 |
| 11 | svmf_rank | 11.08 | asddf_rank | 11.94 | fpgf | 12.75 | sbvdf_rank | 11.21 | svmf_mean | 11.52 | fpgf | 11.25 |
| 12 | fpgf | 12.23 | ffnrf | 12.93 | asddf_rank | 12.89 | fpgf | 11.80 | fpgf | 12.71 | svmf_mean | 12.37 |
| 13 | asddf_rank | 13.33 | fpgf | 13.64 | ffnrf | 13.52 | svmf_mean | 12.90 | asddf_rank | 12.94 | asddf_rank | 14.33 |
| 14 | svmf_mean | 13.35 | asddf_mean | 14.34 | eddf | 14.47 | avmf | 13.32 | mcwvmf | 14.93 | vsdromf | 14.96 |
| 15 | sbvdf_mean | 15.03 | eddf | 15.03 | asddf_mean | 15.16 | asddf_rank | 13.93 | eddf | 16.07 | eddf | 15.98 |
| 16 | avmf | 15.93 | mcwvmf | 15.19 | asvmf_mean | 15.57 | sbvdf_mean | 16.53 | asddf_mean | 16.09 | asvmf_mean | 16.04 |
| 17 | asddf_mean | 16.01 | sbvdf_mean | 17.02 | vsdromf | 15.88 | asddf_mean | 16.78 | evmf | 17.28 | evmf | 17.12 |
| 18 | eddf | 18.06 | evmf | 17.24 | evmf | 16.40 | vsdromf | 18.44 | vsdromf | 17.47 | vmrhf | 17.33 |
| 19 | vsdromf | 18.90 | asvmf_mean | 17.91 | vmrhf | 18.15 | eddf | 18.66 | asvmf_mean | 17.62 | asddf_mean | 17.94 |
| 20 | asvmf_rank | 19.11 | asbvdf_rank | 18.13 | asvmf_rank | 18.82 | asvmf_rank | 18.69 | asvmf_rank | 18.52 | asvmf_rank | 18.64 |
| 21 | asbvdf_rank | 19.12 | vsdromf | 18.31 | asbvdf_rank | 19.09 | evmf | 19.20 | sbvdf_mean | 19.66 | kvmf | 21.07 |
| 22 | evmf | 19.53 | asvmf_rank | 18.89 | sbvdf_mean | 20.86 | asbvdf_rank | 20.01 | asbvdf_rank | 19.97 | asbvdf_rank | 21.28 |
| 23 | asvmf_mean | 20.89 | vmrhf | 22.36 | kvmf | 21.96 | asvmf_mean | 20.57 | avmf | 20.09 | fvmrhf | 21.92 |
| 24 | ebvdf | 23.48 | avmf | 23.09 | fvmrhf | 22.69 | vmrhf | 24.46 | vmrhf | 21.99 | avmf | 23.02 |
| 25 | asbvdf_mean | 23.79 | kvmf | 24.33 | mcwvmf | 23.70 | ebvdf | 24.52 | kvmf | 23.78 | sbvdf_mean | 23.65 |
| 26 | vmrhf | 24.67 | asbvdf_mean | 24.67 | fddrhf | 26.47 | asbvdf_mean | 24.81 | fvmrhf | 25.33 | mcwvmf | 24.16 |
| 27 | kvmf | 25.22 | fvmrhf | 25.48 | asbvdf_mean | 27.11 | kvmf | 24.83 | asbvdf_mean | 26.59 | fddrhf | 26.08 |
| 28 | fvmrhf | 26.80 | ebvdf | 25.69 | avmf | 27.39 | fvmrhf | 26.57 | ebvdf | 27.79 | fvmf | 29.74 |
| 29 | fddrhf | 28.95 | fddrhf | 28.19 | ebvdf | 29.29 | fddrhf | 28.96 | fddrhf | 28.09 | ddf | 30.08 |
| 30 | ddf | 31.37 | ddf | 31.15 | ddf | 30.12 | ddf | 31.50 | ddf | 31.08 | vmf | 30.24 |
| 31 | vmf | 32.80 | fvmf | 31.87 | fvmf | 30.33 | vmf | 32.77 | fvmf | 31.69 | asbvdf_mean | 30.28 |
| 32 | exvmf | 33.08 | vmf | 32.03 | vmf | 30.69 | exvmf | 32.91 | vmf | 31.74 | fovmf | 30.43 |
| 33 | fvmf | 33.11 | fovmf | 32.50 | fovmf | 30.83 | fvmf | 33.01 | fovmf | 32.30 | ebvdf | 32.06 |
| 34 | fovmf | 33.97 | exvmf | 33.21 | exvmf | 32.79 | fovmf | 33.80 | exvmf | 32.78 | exvmf | 32.08 |
| 35 | cbrf | 34.38 | cbrf | 34.27 | ahdf | 33.55 | ahdf | 34.41 | ahdf | 34.18 | ahdf | 33.20 |
| 36 | ahdf | 34.48 | ahdf | 34.40 | cbrf | 33.64 | cbrf | 34.47 | cbrf | 34.47 | cbrf | 33.84 |
| 37 | amnfe | 34.98 | hdf | 35.75 | hdf | 34.98 | amnfe | 35.28 | hdf | 35.42 | hdf | 34.49 |
| 38 | hdf | 35.98 | amnfe | 36.29 | atvmf | 37.16 | hdf | 35.76 | amnfe | 37.34 | atvmf | 36.37 |
| 39 | amnfg | 38.23 | atvmf | 38.69 | amnfe | 38.01 | amnfg | 38.43 | atvmf | 37.84 | amnfe | 39.05 |
| 40 | atvmf | 39.86 | amnfg | 39.14 | fovdf | 39.79 | atvmf | 39.76 | amnfg | 39.74 | fovdf | 39.17 |
| 41 | fovdf | 41.87 | fovdf | 40.96 | amnfg | 39.96 | fovdf | 41.80 | fovdf | 40.70 | gvdf | 40.26 |
| 42 | annmf | 42.49 | gvdf | 41.63 | gvdf | 40.92 | annmf | 42.46 | gvdf | 41.36 | amnfg | 40.57 |
| 43 | gvdf | 42.78 | bvdf | 42.42 | bvdf | 41.27 | gvdf | 42.56 | bvdf | 42.03 | bvdf | 40.71 |
| 44 | bvdf | 43.42 | annmf | 43.88 | fvdf | 44.00 | bvdf | 43.16 | annmf | 43.95 | fvdf | 43.85 |
| 45 | annf | 44.11 | fvdf | 44.79 | annmf | 44.61 | annf | 44.35 | fvdf | 44.37 | annmf | 44.55 |
| 46 | fvdrhf | 45.91 | annf | 45.33 | annf | 45.67 | fvdf | 45.88 | annf | 45.53 | annf | 45.67 |
| 47 | fvdf | 45.95 | fvdrhf | 46.56 | fvdrhf | 46.70 | fvdrhf | 46.06 | fvdrhf | 46.59 | fvdrhf | 46.63 |

**Table 12. Comparison of the filters based on execution time (AR: average ranking)**

| TIME | Uncorrelated Noise | | | | | | Correlated Noise | | | | | |
|---|---|---|---|---|---|---|---|---|---|---|---|---|
| | 5% | | 10% | | 15% | | 5% | | 10% | | 15% | |
| **Rank** | **Filter** | **AR** | **Filter** | **AR** | **Filter** | **AR** | **Filter** | **AR** | **Filter** | **AR** | **Filter** | **AR** |
| 0 | fpgf | 0.08 | fpgf | 0.20 | fpgf | 0.50 | fpgf | 0.06 | fpgf | 0.27 | asvmf_mean | 0.47 |
| 1 | asvmf_mean | 0.92 | asvmf_mean | 0.81 | asvmf_mean | 0.54 | asvmf_mean | 0.94 | asvmf_mean | 0.75 | fpgf | 0.54 |
| 2 | pgf | 2.01 | pgf | 2.02 | pgf | 2.11 | pgf | 2.02 | pgf | 2.07 | pgf | 2.26 |
| 3 | svmf_mean | 3.00 | svmf_mean | 2.97 | svmf_mean | 2.86 | svmf_mean | 2.99 | svmf_mean | 2.91 | svmf_mean | 2.73 |
| 4 | ffnrf | 3.99 | ffnrf | 4.00 | ffnrf | 3.99 | ffnrf | 3.99 | ffnrf | 4.00 | ffnrf | 4.00 |
| 5 | mcwvmf | 5.44 | mcwvmf | 5.33 | mcwvmf | 5.10 | mcwvmf | 5.39 | mcwvmf | 5.29 | mcwvmf | 5.16 |
| 6 | asbvdf_mean | 6.88 | asbvdf_mean | 6.73 | svmf_rank | 7.44 | asbvdf_mean | 7.04 | asbvdf_mean | 7.09 | svmf_rank | 7.32 |
| 7 | svmf_rank | 7.57 | svmf_rank | 7.66 | asbvdf_mean | 7.52 | fmvmf | 7.78 | svmf_rank | 7.41 | asbvdf_mean | 7.61 |
| 8 | fmvmf | 8.06 | fmvmf | 7.71 | fmvmf | 7.72 | svmf_rank | 7.83 | fmvmf | 7.91 | fmvmf | 7.73 |
| 9 | vmf | 8.48 | vmf | 8.47 | vmf | 8.15 | vmf | 8.33 | vmf | 8.21 | vmf | 8.02 |
| 10 | asvmf_rank | 9.85 | asvmf_rank | 9.56 | asvmf_rank | 9.24 | asvmf_rank | 9.74 | asvmf_rank | 9.46 | asvmf_rank | 9.34 |
| 11 | asddf_mean | 10.23 | asddf_mean | 10.85 | asddf_mean | 11.27 | asddf_mean | 10.33 | asddf_mean | 10.92 | asddf_mean | 11.27 |
| 12 | sbvdf_mean | 12.46 | exvmf | 12.38 | exvmf | 11.83 | exvmf | 12.41 | exvmf | 11.89 | exvmf | 11.68 |
| 13 | exvmf | 12.60 | sbvdf_mean | 12.83 | sbvdf_mean | 13.32 | sbvdf_mean | 12.78 | sbvdf_mean | 13.54 | kvmf | 13.95 |
| 14 | kvmf | 14.53 | kvmf | 14.78 | kvmf | 14.80 | kvmf | 14.52 | kvmf | 14.68 | avmf | 14.40 |
| 15 | avmf | 15.05 | avmf | 14.98 | avmf | 14.99 | avmf | 14.86 | avmf | 14.88 | atvmf | 14.65 |
| 16 | atvmf | 15.46 | atvmf | 15.38 | atvmf | 15.18 | atvmf | 15.58 | atvmf | 15.30 | sbvdf_mean | 15.55 |
| 17 | vsdromf | 17.39 | vsdromf | 17.17 | vsdromf | 17.17 | vsdromf | 17.32 | vsdromf | 17.30 | vsdromf | 17.24 |
| 18 | vmrhf | 18.12 | vmrhf | 17.67 | vmrhf | 17.68 | vmrhf | 17.85 | vmrhf | 17.62 | vmrhf | 17.55 |
| 19 | sddf_mean | 19.01 | acwvmf | 19.22 | acwvmf | 19.01 | acwvmf | 19.44 | acwvmf | 19.07 | acwvmf | 19.02 |
| 20 | acwvmf | 19.72 | sddf_mean | 20.17 | sddf_mean | 20.58 | sddf_mean | 19.87 | annmf | 20.71 | annmf | 20.03 |
| 21 | annmf | 21.13 | annmf | 21.02 | annmf | 20.91 | annmf | 21.11 | sddf_mean | 20.92 | amnfe | 22.00 |
| 22 | amnfe | 22.18 | amnfe | 22.36 | amnfe | 22.30 | amnfe | 22.19 | amnfe | 22.37 | cbrf | 22.21 |
| 23 | cbrf | 23.50 | cbrf | 23.35 | cbrf | 23.11 | evmf | 23.24 | cbrf | 23.09 | evmf | 22.84 |
| 24 | evmf | 23.56 | evmf | 23.54 | evmf | 23.88 | cbrf | 23.54 | evmf | 23.39 | sddf_mean | 23.73 |
| 25 | amnfg | 24.33 | amnfg | 24.38 | amnfg | 24.29 | amnfg | 24.44 | amnfg | 24.56 | amnfg | 24.25 |
| 26 | bvdf | 26.00 | bvdf | 25.98 | bvdf | 25.98 | bvdf | 25.92 | bvdf | 25.95 | bvdf | 25.98 |
| 27 | sbvdf_rank | 26.96 | sbvdf_rank | 26.88 | sbvdf_rank | 26.89 | sbvdf_rank | 26.98 | sbvdf_rank | 26.85 | sbvdf_rank | 26.83 |
| 28 | ebvdf | 28.50 | ebvdf | 28.47 | ebvdf | 28.66 | ebvdf | 28.61 | ebvdf | 28.67 | ebvdf | 28.66 |
| 29 | asbvdf_rank | 29.04 | asbvdf_rank | 29.00 | asbvdf_rank | 28.78 | asbvdf_rank | 28.85 | asbvdf_rank | 28.81 | asbvdf_rank | 28.82 |
| 30 | gvdf | 29.47 | gvdf | 29.51 | gvdf | 29.56 | gvdf | 29.59 | gvdf | 29.52 | gvdf | 29.51 |
| 31 | abvdf | 31.04 | abvdf | 31.00 | abvdf | 30.95 | abvdf | 31.01 | abvdf | 31.02 | abvdf | 30.96 |
| 32 | eddf | 32.46 | eddf | 32.47 | annf | 32.68 | eddf | 32.47 | eddf | 32.61 | annf | 32.48 |
| 33 | annf | 32.81 | annf | 32.80 | eddf | 32.98 | annf | 32.88 | annf | 32.78 | eddf | 33.13 |
| 34 | ddf | 33.58 | ddf | 33.65 | ddf | 33.28 | ddf | 33.48 | ddf | 33.49 | ddf | 33.38 |
| 35 | sddf_rank | 34.98 | sddf_rank | 35.05 | sddf_rank | 35.05 | sddf_rank | 34.93 | sddf_rank | 34.93 | sddf_rank | 34.87 |
| 36 | acwvdf | 35.96 | acwvdf | 35.95 | acwvdf | 36.05 | acwvdf | 36.07 | acwvdf | 36.25 | acwvdf | 36.34 |
| 37 | asddf_rank | 36.85 | asddf_rank | 36.88 | asddf_rank | 36.83 | asddf_rank | 36.82 | asddf_rank | 36.72 | asddf_rank | 36.68 |
| 38 | hdf | 37.94 | hdf | 37.94 | hdf | 37.94 | hdf | 37.94 | hdf | 37.94 | hdf | 37.94 |
| 39 | ahdf | 39.10 | ahdf | 39.01 | ahdf | 38.99 | ahdf | 39.03 | ahdf | 38.96 | ahdf | 38.98 |
| 40 | acwddf | 39.93 | acwddf | 40.01 | acwddf | 40.01 | acwddf | 40.01 | acwddf | 40.07 | acwddf | 40.05 |
| 41 | fvdf | 40.93 | fvdf | 40.95 | fvdf | 40.96 | fvdf | 40.92 | fvdf | 40.92 | fvdf | 40.94 |
| 42 | fovdf | 41.96 | fovdf | 41.96 | fovdf | 41.96 | fovdf | 41.96 | fovdf | 41.96 | fovdf | 41.96 |
| 43 | fvmrhf | 43.01 | fvmrhf | 43.01 | fvmrhf | 43.02 | fvmrhf | 43.01 | fvmrhf | 43.01 | fvmrhf | 43.01 |
| 44 | fvdrhf | 43.96 | fvdrhf | 43.96 | fvdrhf | 43.96 | fvdrhf | 43.96 | fvdrhf | 43.96 | fvdrhf | 43.96 |
| 45 | fddrhf | 45.03 | fddrhf | 45.03 | fddrhf | 45.00 | fddrhf | 45.03 | fddrhf | 45.02 | fddrhf | 45.00 |
| 46 | fvmf | 45.97 | fvmf | 45.97 | fvmf | 45.99 | fvmf | 45.97 | fvmf | 45.97 | fvmf | 45.98 |
| 47 | fovmf | 46.97 | fovmf | 46.98 | fovmf | 46.99 | fovmf | 46.97 | fovmf | 46.98 | fovmf | 46.99 |

**Table 13. Most effective filters at each noise level**

| Noise Level | Most Effective Filters |
|---|---|
| 5% Noise | *ACWDDF*, *PGF*, *SDDF_rank*, *ACWVMF*, *ACWVDF*, *MCWVMF*, *SDDF_mean* |
| 10% Noise | *ACWDDF*, *PGF*, *SDDF_rank*, *ACWVMF*, *SDDF_mean*, *SVMF_rank* |
| 15% Noise | *ACWDDF*, *PGF*, *SDDF_rank*, *ACWVMF*, *SVMF_rank* |

FIGURE LEGEND

Figure 1. Representative images from the image set

(a) flowerbee

(b) cat

(c) Austria

(d) Scotland

(e) Capilano Suspension Bridge

(f) Native American

(g) sweetgum

(h) dermoscopy

(i) fractal

Figure 2. Sample filtering results for the baboon image

(a) Original

(b) 10% correlated noise

MAE: 6.058; MSE: 893.707; NCD: 0.101

(c) ACWDDF

MAE: 1.902; MSE: 76.956; NCD: 0.012

(d) ACWVDF

MAE: 2.182; MSE: 102.892; NCD: 0.014

(e) PGF

MAE: 2.293; MSE: 98.825; NCD: 0.015

(f) SDDF_mean

MAE: 3.017; MSE: 124.358; NCD: 0.019

(g) SDDF_rank

MAE: 3.031; MSE: 123.020; NCD: 0.019

(h) ACWVMF

MAE: 3.726; MSE: 171.007; NCD: 0.023

Figure 3. Sample filtering results for the Native American image

(a) Original

(b) 15% correlated noise

MAE: 9.600; MSE: 1558.290; NCD: 0.182

(c) ACWDDF

MAE: 1.453; MSE: 49.316; NCD: 0.015

(d) PGF

MAE: 1.593; MSE: 54.189; NCD: 0.021

(e) SDDF_rank

MAE: 1.594; MSE: 50.284; NCD: 0.016

(f) ACWVMF

MAE: 1.643; MSE: 53.992; NCD: 0.019

(g) SDDF_mean

MAE: 1.776; MSE: 74.073; NCD: 0.021

(h) ACWVDF

MAE: 2.030; MSE: 109.360; NCD: 0.018

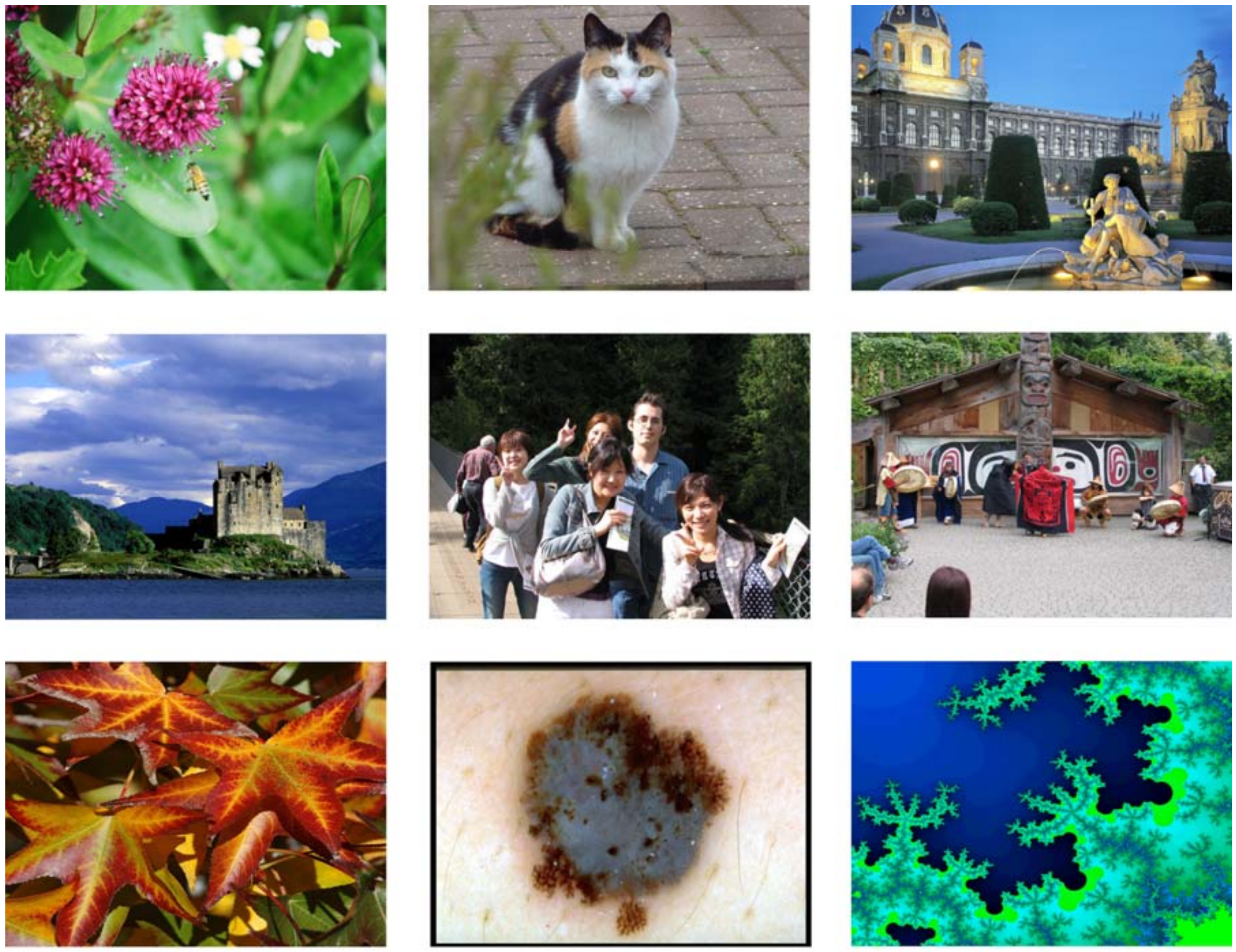

**Figure 1.** Representative images from the image set

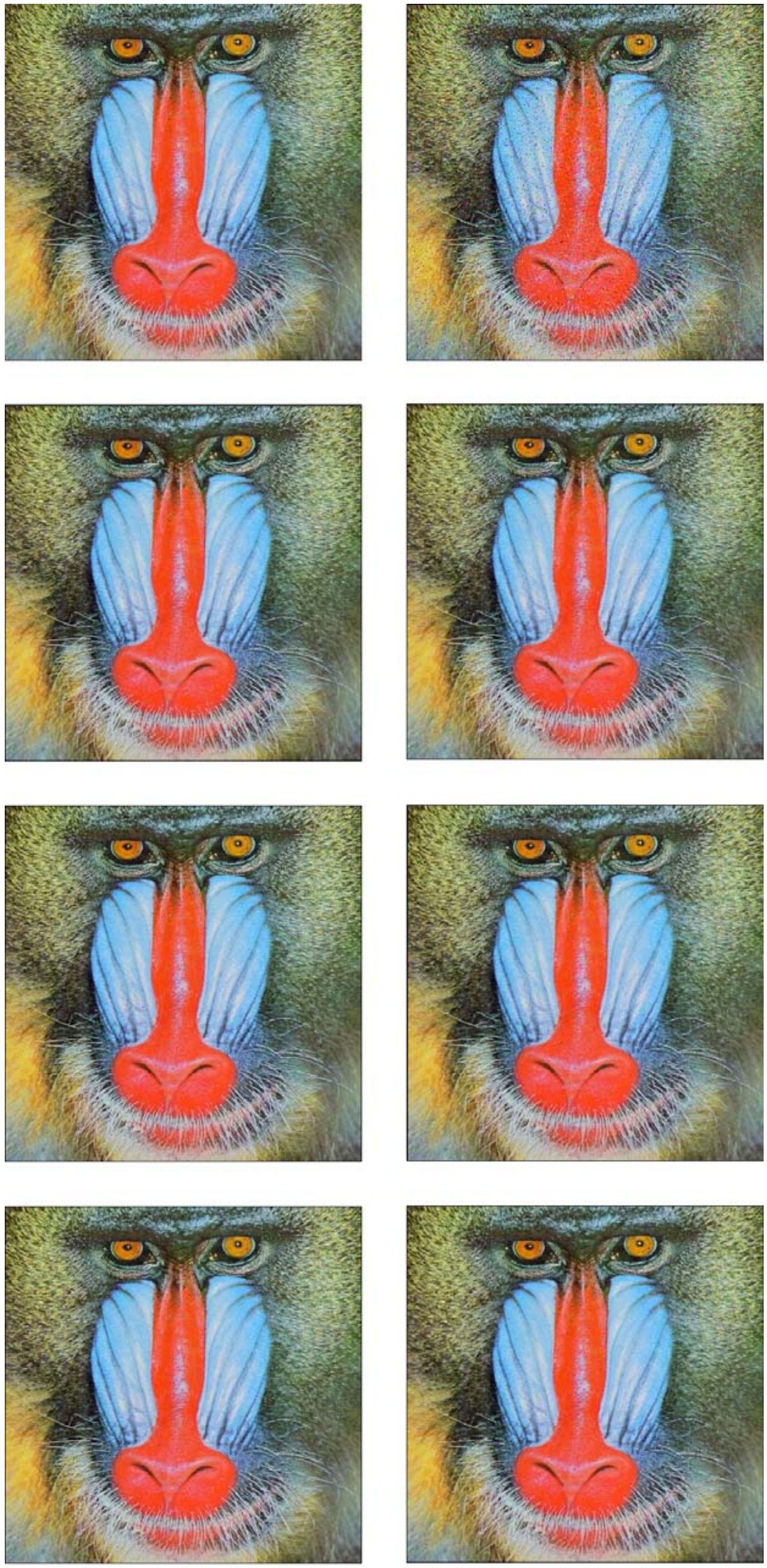

**Figure 2.** Sample filtering results for the baboon image

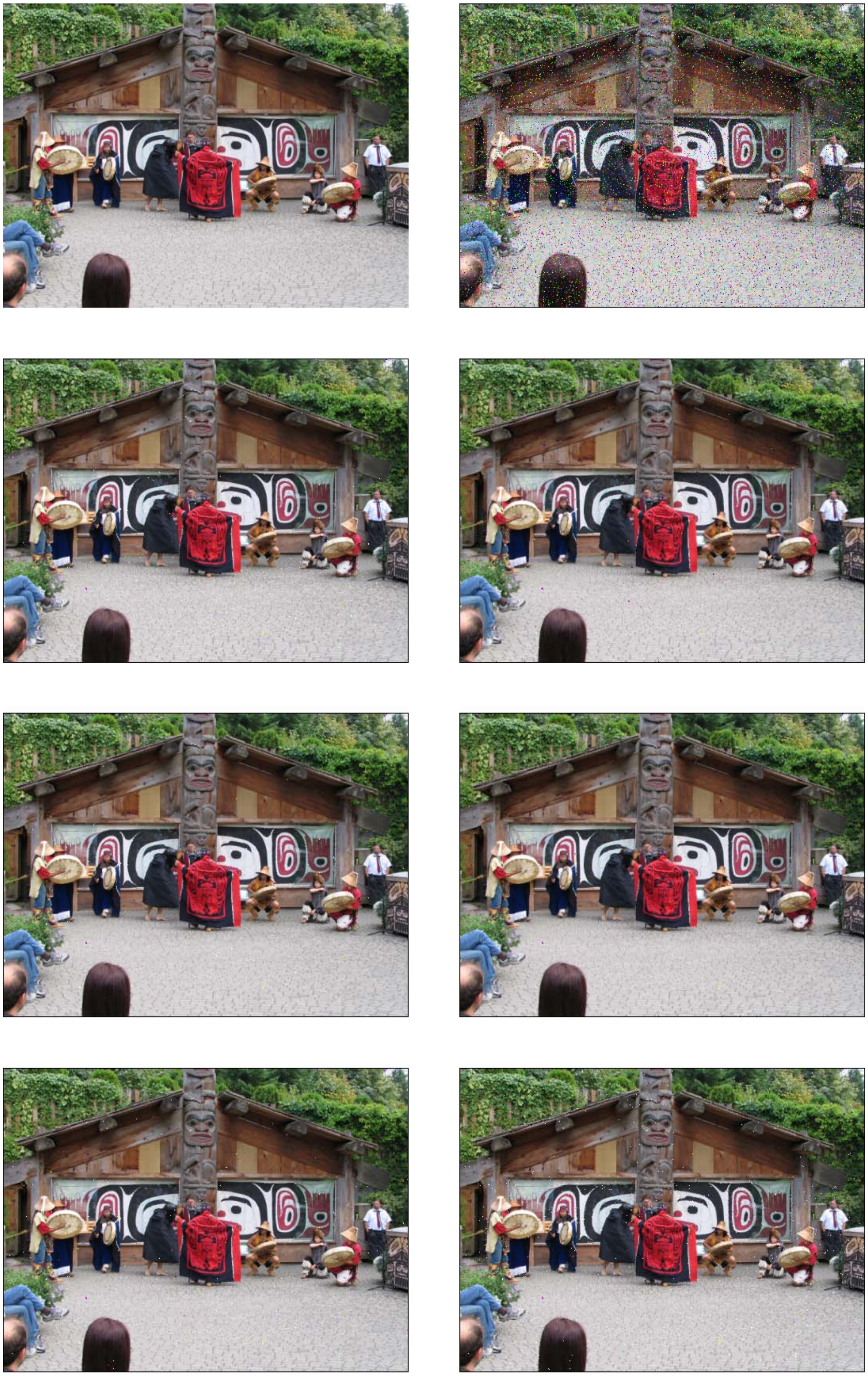

**Figure 3.** Sample filtering results for the Native American image